\documentclass[10pt,twocolumn,letterpaper]{article}

\usepackage[pagenumbers]{iccv} 

\definecolor{iccvblue}{rgb}{0.21,0.49,0.74}
\usepackage[pagebackref,breaklinks,colorlinks,allcolors=iccvblue]{hyperref}

\usepackage{lipsum}
\usepackage{moresize}
\usepackage{multirow}
\usepackage[capitalize]{cleveref}
\usepackage{wrapfig}
\usepackage{amsmath}
\usepackage{amssymb}
\usepackage{booktabs}
\usepackage{cases}
\usepackage{algorithm}
\usepackage{algpseudocode}
\usepackage{xcolor}
\newcommand{\colororange}[1]{%
  \colorbox{orange!15
  }{$\displaystyle#1$}}
\newcommand{\colorblue}[1]{%
  \colorbox{blue!8
  }{$\displaystyle#1$}}

\usepackage{overpic}
\usepackage{rotating}
\usepackage{cancel}
\usepackage{pgfplotstable} 
\usepackage{tikz}
\usepackage{tikz-cd}
\usepackage{colortbl}
\usepackage{cancel}
\usepackage{multibib}
\usepackage{afterpage} 

\makeatletter
\AddToHook{cmd/appendix/before}{\def\cref@section@alias{appendix}}
\makeatother

\crefname{section}{Sec.}{Secs.}
\Crefname{section}{Section}{Sections}
\Crefname{table}{Table}{Tables}
\crefname{table}{Tab.}{Tabs.}
\Crefname{append}{Appendix}{Appendixs}
\crefname{append}{Append.}{Appends.}
\Crefname{subfigure}{Figure}{Figures}
\crefname{subfigure}{Fig.}{Figs.}

\newcommand{\mname}{{Di$\mask{}$O}}

\newcommand{\fakemodel}{auxiliary model}
\newcommand{\mask}{\mathtt{[M]}}
\newcommand{\method}{{Di$\mask{}$O}}

\def\paperID{8824} 
\def\confName{ICCV}
\def\confYear{2025}

\title{\mname{}: Distilling Masked Diffusion Models into One-step Generator}

\author{Yuanzhi Zhu$^{1}$
\quad
Xi Wang$^1$
\quad
Stéphane Lathuilière$^{2}$
\quad
Vicky Kalogeiton$^{1}$\\
$^1$LIX, École Polytechnique, CNRS, IPP \quad
$^2$Inria, Univ. Grenoble Alpes, CNRS, LJK\\
\href{https://yuanzhi-zhu.github.io/DiMO/}{https://yuanzhi-zhu.github.io/DiMO/}
}

\newcommand{\yzc}[1]{{#1}}
\newcommand{\xwc}[1]{{#1}}

\begin{document}

\twocolumn[{%
 \renewcommand\twocolumn[1][]{#1}
 \maketitle
 \centering
    \vspace{3mm}
    \begin{overpic}[width=1.\linewidth]{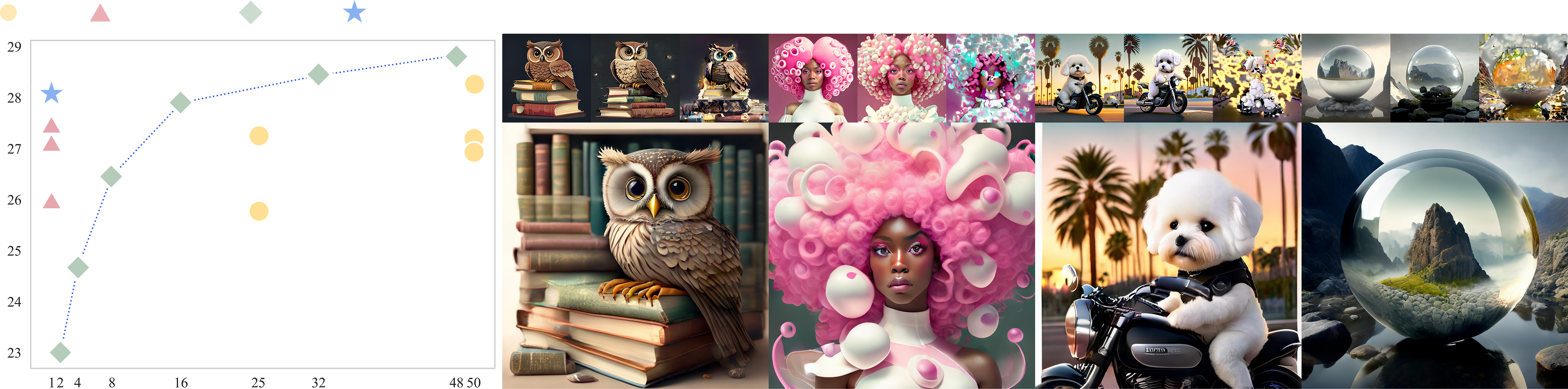}
    \put(59.,-1.5){\color{black}{\scriptsize \textbf{Our One-step Generator}}}
    \put(52.8,23.6){\color{black}{\scriptsize \textit{Teacher}: \textbf{Meissonic}~\cite{bai2024meissonic} (64, 16, 4 steps)}}
    \put(11.6,-1.5){\color{black}{\scriptsize Generation Steps}}
    \put(-1.,6.5){\color{black}{\scriptsize \rotatebox{90}{HPSv2~\cite{wu2023human} Score}}}
    \put(1.8, 23.4){\color{black}{\tiny DMs}}
    \put(7.5, 23.4){\color{black}{\tiny One-step DMs}}
    \put(17.2,23.4){\color{black}{\tiny MDM}}
    \put(23.7,23.4){\color{black}{\tiny \textbf{One-step MDM}}}
    \put(4.4, 19.1){\color{black}{\scriptsize \textbf{\mname{}}}}
    \put(4.4, 17.8){\color{black}{\scriptsize (\textit{ours})}}
    \put(11., 20.7){\color{black}{\scriptsize \textbf{Meissonic ICLR'25~\cite{bai2024meissonic}}}}
    \put(11., 19.5){\color{black}{\scriptsize (\textit{teacher})}}
    \put(4.2, 16.5){\color{gray}{\tiny SD Turbo~\cite{sauer2024adversarial}}}
    \put(4.2, 15.2){\color{gray}{\tiny SwiftBrush v2~\cite{dao2025swiftbrush}}}
    \put(4.2, 11.6){\color{gray}{\tiny InstaFlow~\cite{liu2023instaflow}}}
    \put(17.7, 16.2){\color{gray}{\tiny Latent}}
    \put(17.7, 15.2){\color{gray}{\tiny Diffusion~\cite{rombach2022high}}}
    \put(17.7, 11.0){\color{gray}{\tiny DeepFloyd-XL~\cite{deepfloyddeepfloyd}}}
    \put(26.5, 19.3){\color{gray}{\tiny SDXL}}
    \put(22.1, 18.4){\color{gray}{\tiny Base/Refiner~\cite{rombach2022high}}}
    \put(24.2, 15.7){\color{gray}{\tiny SDv2.0~\cite{rombach2022high}}}
    \put(24.2, 14.7){\color{gray}{\tiny SDv1.4~\cite{rombach2022high}}}
    \end{overpic}
    \captionof{figure}{
    Unlike \textcolor[RGB]{249, 194, 64}{continuous Diffusion Models (DMs)} that \yzc{have beed} successfully distilled into \textcolor[RGB]{229, 166, 173}{one-step DM generators} with performances competitive with the teacher, 
    distilling \textcolor[RGB]{180, 204, 185}{Masked Diffusion Models (MDM)} into one-step generator remains a challenge.  
    \yzc{In this paper, we propose the \textbf{first}}
    \textcolor[RGB]{137, 177, 242}{one-step distillation method: \mname{} for MDM}, 
    e.g., from the recent text-to-image masked diffusion model Meissonic~\cite{bai2024meissonic}. We demonstrate that our \mname{} can successfully distill the teacher model into a one-step generator, achieving competitive performance both quantitatively and qualitatively, while the teacher model's performance deteriorates rapidly with reduced generation steps, i.e., comparing our one-step results (large, bottom images) to 4-step teacher outputs (right corner of each image).
    }
    \label{fig:teaser}
    \vspace{5mm}
}]
\maketitle

\begin{abstract}
Masked Diffusion Models (MDMs) have emerged as a powerful generative modeling technique. Despite their remarkable results, they typically suffer from slow inference with several steps. In this paper, we propose \mname{}, a novel approach that distills masked diffusion models into a one-step generator. 
\mname{} addresses two key challenges: (1) the intractability of using intermediate-step information for one-step generation, which we solve through token-level distribution matching that optimizes model output logits by an `on-policy framework' with the help of an auxiliary model; and (2) the lack of entropy in the initial distribution, which we address through a token initialization strategy that injects randomness while maintaining similarity to teacher training distribution. 
We show \mname{}'s effectiveness on both class-conditional and text-conditional image generation, impressively achieving performance competitive to multi-step teacher outputs while drastically reducing inference time. To our knowledge, we are the first to successfully achieve one-step distillation of masked diffusion models and the first to apply discrete distillation to text-to-image generation, opening new paths for efficient generative modeling. 
\end{abstract}

\section{Introduction}
Recently, Masked Diffusion Models (MDMs) \cite{austin2021structured,chang2022maskgit,shi2025simplified,hu2024mask} have emerged as a prominent framework in generative modeling, demonstrating strong performance in a broad spectrum of tasks. 
Unlike auto-regressive models that generate data sequentially, MDMs enable faster discrete data generation. 
Moreover, compared to continuous diffusion models~\cite{ho2020denoising}, MDMs are particularly advantageous for visual modeling within multimodal frameworks, as they leverage a unified vocabulary to seamlessly integrate visual and textual content.
Their versatility has enabled successful applications in image generation \cite{chang2022maskgit,chang2023muse,hu2024mask,bai2024meissonic,patil2024amused,issenhuth2021edibert}, text generation \cite{nie2024scaling,nie2025large,loudiscrete,gong2024scaling,deschenaux2024beyond}, video generation \cite{susladkar2024motionaura,fuest2025maskflow}, multimodal modeling \cite{hu2022unified,li2024dual,xie2024show,wang2024dplm}, protein design \cite{gruver2023protein,campbell2024generative}, audio synthesis~\cite{sadok2025ancogen}, and motion generation \cite{chi2024m2d2m,pinyoanuntapong2024mmm}. In addition, several algorithms have been proposed to guide the generation process in MDMs \cite{li2024derivative,schiff2024simple,rector2024steering,singhal2025general,nisonoff2024unlocking}, further enhancing their flexibility and alignment with user-defined objectives.

Despite these advantages, current MDMs suffer from low inference speed, i.e. high number of sampling steps required during generation~\cite{feng2025theoretical}.
Besides accelerated sampling techniques \cite{zhao2024informed,ren2025fast,campbell2022continuous,park2024optimizing,kim2025train,besnier2025halton} that yield good performances but tend to drastically deteriorate the generation with fewer inference steps, distillation techniques address this efficiency bottleneck. 
They transfer knowledge from a well-performing but slow teacher model to a student model capable of achieving comparable results with significantly fewer steps (even one). 
For this reason, distillation has been widely adopted for continuous diffusion models, typically with consistency-based~\cite {song2023consistency,song2023improved,lu2024simplifying} or distillation matching approaches~\cite{yin2024one,xu2024ufogen,luo2023diff}. These effectively reduce inference time while maintaining generation quality. However, they cannot be directly applied to masked diffusion due to fundamental differences in their mathematical formulation, i.e. the absence of a Probability Flow Ordinary Differential Equation (PF-ODE) and the inability to re-parameterize MDM outputs as \yzc{denoising} score functions {\cite{meng2022concrete,loudiscrete,deschenaux2024beyond}.}
\yzc{Furthermore, recent work \cite{song2025ideas} points out that multi-token prediction in MDMs is inherently challenging due to their na\"ive conditional independence assumption.}

To overcome this, the community is just starting to explore distillation alternatives for masked diffusion \cite{hayakawa2024distillation,deschenaux2024beyond}.
\yzc{
\cite{hayakawa2024distillation,deschenaux2024beyond} are the only works attempting this. Hayakawa \etal \cite{hayakawa2024distillation} propose viewing MDMs as mixture models, progressively distilling dimension correlations into a few-step student. Meanwhile, Deschenaux \etal \cite{deschenaux2024beyond} collect the teacher's multi-step output log probability as a target for the student to learn with fewer steps. Despite their improved efficiency, these methods have critical limitations: both require a multi-round distillation process, which is computationally expensive and prone to error accumulation when further reducing student steps.
}

Instead, in this work, \yzc{we directly address \textbf{one-step distillation for masked diffusion models}.} While achieving this, we identified two main challenges. 
First, there is the difficulty of leveraging the intermediate steps information for one-step generator 
\footnote{\yzc{Without explicit distinction, we use the terms ``\textit{one-step generator}" and ``\textit{student model}" interchangeably when referring to our distilled model.}}.
To distill the teacher's whole multi-step information into a \textit{one-step} model, neither can we compute the loss between the teacher's intermediate outputs and intractable student intermediate ones, nor can we update the one-step generator using the loss gradients due to the non-differentiable sampling operation. 
Second, the lack of entropy in the initial distribution used during one-step generation \cite{hayakawa2024distillation}. While continuous diffusion models typically start with a Gaussian distribution that naturally incorporates randomness, MDM begins with a sequence of identical mask tokens. This lack of entropy can lead to mode collapse in one-step distillation, making effective distribution matching more difficult.

Therefore, in this work, we propose \textbf{Di}stilling $\mathtt{[M]}$asked diffusion models into a \textbf{O}ne-step generator, coined \mname{}, trained to transform an initial input distribution into outputs that closely resemble the teacher’s multi-step generation distribution. Inspired by \textit{data-free} on-policy distillation~\cite{agarwal2024policy,ross2011reduction,ho2016model,balakrishna2020policy,arora2022exposure,schmidt2019generalization,hinton2015distilling}, \mname{} matches the conditional \textit{token-level} distribution along all intermediate states to ensure the overall generation distribution matching. 
Our first technical contribution is that \mname{} approximates the training objective using pseudo-intermediate states combined with an auxiliary model that surrogates the gradient.
Our second technical contribution addresses the insufficient entropy in the initial state of one-step generation. \mname{} employs a token initialization strategy that injects randomness while maintaining similarity to the input sequences used to train the teacher, ensuring robust distillation while effectively avoiding mode collapse. 

Using image generation as a representative example, we conduct experiments on class-conditional generation with MaskGit~\cite{chang2022maskgit}, and on text-conditional image generation with Meissonic~\cite{bai2024meissonic} as the teacher model. Our results show that for both tasks, \mname{} can efficiently distill the teacher model into a one-step generator while impressively achieving performance competitive with the multi-step outputs of the teachers. Our findings also contain training objective with generalized divergence, ablation studies discussing our design choices, limitations and future directions.

Our contributions can be summarized as:
(1) We are the first to successfully achieve one-step distillation of masked diffusion models, paving the path for future work.  
(2) To address this, we propose \mname{}, a token-level distillation method that enables the one-step distillation of masked diffusion models, with proposed efficient token initialization. 
(3) Our findings show that \mname{} 
successfully reaches performance close to that of MDM teachers, while \yzc{greatly enhances the sampling efficiency}, verified by both class-conditional and text-conditional image generation tasks; note that, \yzc{\mname{} is} the first to experiment with discrete \textit{distillation} for text-to-image generation. 

\section{Related Works}
\label{sec:relatedworks}

\noindent\textbf{Acceleration of Masked Diffusion Models.}
Improving the efficiency of MDMs has been an active area of research, with various methods proposed to accelerate inference while maintaining generation quality \cite{zhao2024informed,ren2025fast,campbell2022continuous,park2024optimizing,kim2025train,besnier2025halton}.
For instance, Park \etal \cite{park2024optimizing} minimize compounding decoding error to improve the allocation of discrete sampling timesteps, achieving efficiency gains without additional computational overhead.  
Ren \etal \cite{ren2025fast} introduce a higher-order solver that allows for larger step sizes while reducing numerical errors, enabling faster generation with minimal quality loss. 
Other works have explored alternative strategies to enhance efficiency. Hayakawa \etal \cite{hayakawa2024distillation} propose a distillation approach in which a few-step student is trained using a combination of distillation loss and consistency loss to improve sampling efficiency. However, their method relies on either simulating the teacher's trajectory or utilizing real data to obtain intermediate states for training the student, leading to a mismatch between training and inference. 
\yzc{Similar to Progressive Distillation \cite{salimans2022progressive}, Deschenaux \etal \cite{deschenaux2024beyond} distill a student model to approximate the teacher’s multi-step output distribution via divergence minimization. However, this approach requires multiple inferences of the teacher model during student training. Additionally, since the teacher’s output tokens are independent, the distillation step size is constrained, necessitating a multi-round distillation process similar to \cite{salimans2022progressive}.}
Furthermore, all these approaches still require multiple sampling steps and suffer from significant performance degradation when the number of steps is further reduced.
This limitation highlights the need for further research into efficient one-step generation strategies, which is the task we propose and tackle with our proposed \mname{}.

\vspace{0.2cm}
\noindent\textbf{Continuous Diffusion Distillation.}
The goal of diffusion distillation is to reduce sampling steps, and can be generally divided into through two primary approaches \cite{zhu2024accelerating}: regression-based and distribution-based methods \cite{zhang2025towards}. 
Regression-based techniques~\cite{luhman2021knowledge,song2023consistency,liu2022rectified,salimans2022progressive,gu2023boot,meng2023distillation,yan2024perflow,zhu2025slimflow} train student models via regression objectives derived from the PF-ODE of the teacher model.
It is challenging to apply these techniques for MDMs due to their missing of explicit \yzc{PF-ODE as} sampling trajectory.
Distribution-based methods~\cite{luo2023diff,yin2024one,yin2024improved,xu2024ufogen,kim2023consistency,zhou2024score,zhou2024long,zhou2024adversarial,nguyen2024swiftbrush,dao2025swiftbrush,luo2025one} instead directly match the student’s output distribution to the teacher’s multi-step sampling distribution, often leveraging adversarial or divergence-minimization strategies.
Similar to prior Variational Score Distillation (VSD) approaches~\cite{wang2024prolificdreamer,yin2024one,yin2024improved,nguyen2024swiftbrush,dao2025swiftbrush,luo2023diff,xie2024distillation,salimans2025multistep}, our \mname{} approach employs an \fakemodel{} $\psi$ to approximate the intermediate distributions of an one-step generator.
However, our method differs in three key aspects.  
First, it is specifically designed for masked diffusion and does not rely on the score function used in continuous diffusion models. 
Second, while VSD-based methods minimize the \yzc{divergence of} \textit{denoising distributions} \cite{zhang2020spread,he2024training} over the full space of latent images, we instead optimize the \yzc{divergence of} \textit{token-level conditional distributions}.
Third, our framework generalizes beyond the reverse \yzc{Kullback–Leibler (KL)} divergence employed in these prior works, supporting {arbitrary} $f$-divergences \cite{csiszar2004information} for distribution matching, effectively mitigating the mode-seeking bias.

\newcommand{\m}{\mathbf{m}}

\section{Preliminary: Masked Diffusion Models}
\label{sec:background}

Masked Diffusion Models (MDMs) \cite{chang2022maskgit,shi2025simplified,hu2024mask} rely on a pre-trained discrete image tokenizer, such as VQ-VAE~\cite{van2017neural,esser2021taming}. 
Given an image of shape $3 {\times} H {\times} W$, the tokenizer encodes it into a sequence $x_0 = \{ x_{\theta}^i \}_{i=1}^L$ of discrete tokens, where $x_{\theta}^i{\in}[V]$ is the token at position $i$.
Here, $[V]= \{0, 1, \dots, V{-}1\}$ represents the set of token indices, with $V$ the vocabulary size.
The sequence length is given by $L = h w$, where $h$ and $w$ correspond to the height and width of the 2D token grid produced by the VQ-VAE encoder.
Similar to continuous diffusion models \cite{sohl2015deep,song2020score,ho2020denoising}, MDMs generate discrete data by learning to reverse an absorbing-state forward diffusion process \cite{austin2021structured}.

\begin{figure}
    \centering
    \begin{overpic}[width=1.\linewidth]{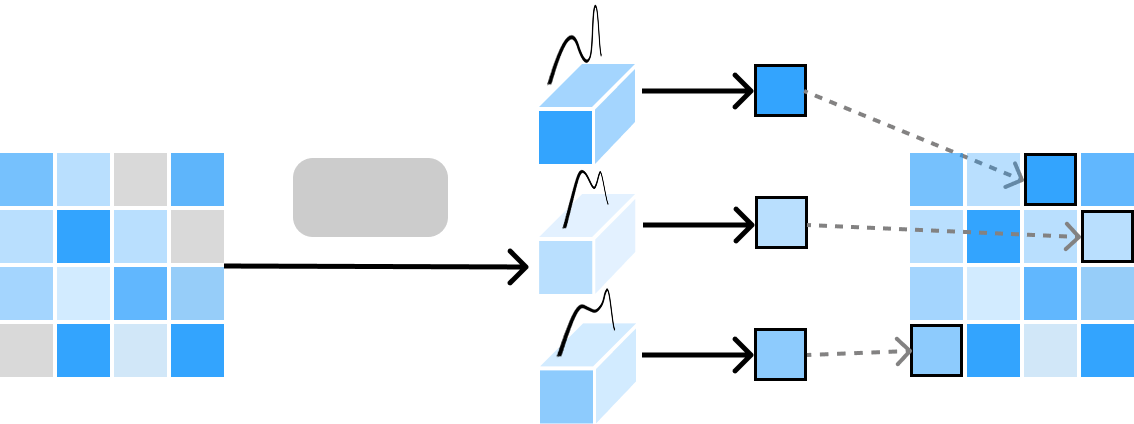}
    \put(8, 1.5){\color{black}{$x_t$}}
    \put(87.9, 1.5){\color{black}{$x_s$}}
    \put(77.9, 28.5){\scriptsize\color{black}{$p\!=\!(r_t\!-\!r_s)/r_t$}}
    \put(66.9, -.4){\color{black}{$x_\phi^i$}}
    \put(48.4, -3.8){\color{black}{$z_\phi^i$}}
    \put(28.9,18.8){\color{black}{\scriptsize MDM}}
    \put(24.6,10.4){\color{black}{\scriptsize Predict Logits}}
    \put(24.9,7.5){\color{black}{\scriptsize \textbf{\textit{independently}}}}
    \end{overpic}
    \vspace{0.01cm}
    \caption{\textbf{Reverse process of MDM.}
    With a masked sequence $x_t$ as input, the MDM \emph{independently} output logits $z_\phi^i$ at each masked position $i$, which are then used to sample the new tokens $x_\phi^i$ as the model prediction. In the next state $x_s$, 
    we use $x_\phi^i$ to replace each masked token in $x_t$ with probability $(r_t-r_s)/r_t$.}
    \label{fig:reverse_process}
    \vspace{-0.2cm}
\end{figure}
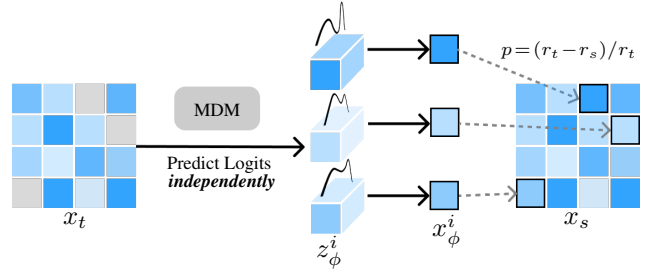

\vspace{0.2cm}
\noindent\textbf{Forward Process.} 
The forward process gradually corrupts the original token sequence $x_0$ into a {fully} masked sequence $x_1$.
For any timestep $t \in [0,1]$, this yields intermediate masked token \textit{sequences} $x_t$, composed of mixed image and mask tokens at a mask ratio $r_t$ \footnote{
In masked diffusion, each token in a sequence can only be either $\mask{}$ token or image token, and different timesteps influence only the proportion of these two token types in the sequence (following the mask ratio $r_t$), e.g. in half timesteps ($r_t=0.5$), half the tokens will be masked.
}.
The transition from $x_0$ to $x_t$ follows the probability~\cite{sahoo2024simple,ou2024your,shi2025simplified}:
\begin{align}
    q_{t|0}(x_t|x_0) &= \prod_{i=0}^{L-1}\text{Cat}(x_t^i;(1-r_t)\delta(x_0^i) + r_t\delta(\mask{})),
\end{align}
where $i \in \{0, 1, \dots, L-1 \}$
denotes the token position, $\mask{}$ specifies the mask token, $\delta(\cdot)$ is the Dirac function, and $\text{Cat}(x^i;p)$ refers to the categorical distribution over $V{+}1$ categories in $[V] \cup \{\mask{}\}$.
Essentially, tokens in $x_0$ are \textit{independently} masked with probability $r_t$ at timestep $t$.

\paragraph{Parameterized Reverse Process.}  
The reverse process reconstructs the original image tokens by iteratively replacing masked tokens with predicted ones, starting from a fully masked sequence $x_1$. 
This process is guided by a model $\phi$, which predicts a categorical distribution over each masked token at position $i$ given $x_t$:  
\begin{align}\label{eq:softmax}
p_\phi(x_0^i | x_t) := \text{softmax}\left({z_\phi^i(x_t)}/{\tau}\right), 
\end{align} 
where $z_\phi^i \in \mathbb{R}^{V}$ \yzc{are} the \yzc{independently predicted} logits, and $\tau$ controls sampling diversity.
As shown in \cref{fig:reverse_process}, each predicted image token $x_\phi^i$ is \emph{independently} sampled from the token-level predicted distribution $p_\phi(x_0^i | x_t)$ and is used to transition from timestep $t$ to $s$ ($0 \leq s < t \leq 1$).
During this transition, {unmasked tokens remain unchanged}, whereas each masked token $x_t^i$ is updated to $x_\phi^i$ with probability $ (r_t-r_s)/r_t$ {or} stay {masked} with probability $r_s/r_t$. The model $\phi$ operates {independently} across token positions, enabling parallel token prediction, hence efficiently refining $x_1$ into $x_0$ in a progressive manner.

\vspace{0.2cm}
\noindent\textbf{MDM Training Objective.}  
The model $\phi$ is trained to reconstruct the original token sequence $x_0$ from their masked counterparts $x_t$ by minimizing the negative log-likelihood across all mask ratios $r_t$:  
\begin{align}
\label{eq:MDM_loss}
    \mathcal{L}_\text{MDM} = \mathbb{E}_{x_0, t}\left[\left(\mathbb{E}_{q_{t|0}} [- \log p_{0|t}(x_0|x_t,\phi)] \right)\right],
\end{align}
with $x_0 {\sim} p_0(x_0)$ and $t {\sim} \mathcal{U}[0,1]$.
This loss is computed as cross-entropy per masked token and is the same objective used in Masked Image Modeling (MIM) approaches \cite{bao2021beit,hu2024mask} or masked generative models \cite{ghazvininejad2019mask,hur2024unlocking}, such as MaskGit~\cite{chang2022maskgit} and Meissonic \cite{bai2024meissonic}, where the models also take condition $c$ (class label or text prompt) as input.

\begin{figure*}[t!]
\vspace{-0.3cm}
\centering
\begin{overpic}[width=1.\linewidth]{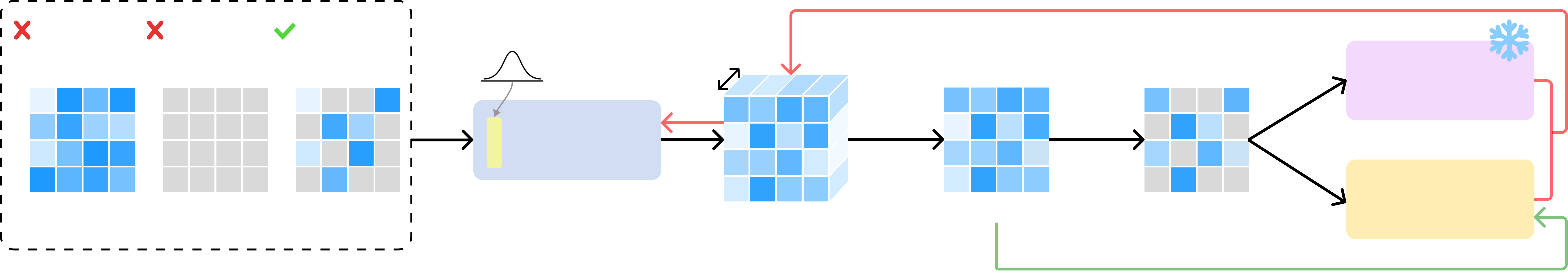}
\put(1.,3.1){\color{black}{\scriptsize Token Initialization Strategy}}
\put(2.2,15.3){\color{black}{\scriptsize{All random}}}
\put(2.9,12.7){\color{black}{\scriptsize{$r_{\text{init}}\!=\!0$}}}
\put(10.6, 15.3){\color{black}{\scriptsize All masked}}
\put(11.7,12.7){\color{black}{\scriptsize{$r_{\text{init}}\!=\!1$}}}
\put(19.2,15.3){\color{black}{\scriptsize Proposed}}
\put(18.5,12.7){\color{black}{\scriptsize{$0\!<\!r_{\text{init}}\!<\!1$}}}
\put(20.9,4.6){\color{black}{\scriptsize $x_{\text{init}}$}}
\put(28.8,15.3){\color{black}{\scriptsize $\mathcal{N}(0,\sigma_{\text{init}}{I})$}}
\put(33.8,8.2){\color{black}{\scriptsize Student $\theta$}}
\put(30.7,4.8){\color{black}{\scriptsize One-step Generator}}
\put(44.8,12.9){\color{black}{\scriptsize $V$}}
\put(50.8,15.2){\color{black}{\scriptsize ${\nabla_{z} {D}(p_\phi||p_\psi)(\tilde{x}_t)}$}}
\put(54.7,9.3){\color{black}{\scriptsize Sample}}
\put(67.4,9.3){\color{black}{\scriptsize Forward}}
\put(64.5,1.25){\color{black}{\scriptsize Cross Entropy}}
\put(46.8,3.8){\color{black}{\scriptsize Logits $z_\theta$}}
\put(60.7,4.3){\color{black}{\scriptsize Tokens $x_\theta$}}
\put(71.6,4.3){\color{black}{\scriptsize Masked Tokens $\tilde{x}_t$}}
\put(88.95,12.1){\color{black}{\scriptsize Teacher $\phi$}}
\put(86.35,4.5){\color{black}{\scriptsize Auxiliary Model $\psi$}}
\end{overpic}
\vspace{-0.5cm}
\caption{\textbf{\mname{} Pipeline.}
Our method distills a costly multi-step MDM teacher into a one-step generator.
Given $x_{\text{init}}$ sampled using our proposed token initialization strategy, the one-step generator ({student $\theta$}) produces logits $z_\theta$, from which image token sequence $x_{\theta}=\{x_{\theta}^i\}_{i=1}^{L}$ are sampled.
These tokens are then processed to obtain an intermediate state $\tilde{x}_t$ through a forward mask diffusion process. 
For each intermediate state $\tilde{x}_t$, we update the one-step generator $\theta$ and \fakemodel{} $\psi$ alternately: the one-step generator is optimized by minimizing the conditional divergence ${{D}(p_\phi||p_\psi)(\tilde{x}_t)}$ at \emph{token-level}, while the \fakemodel{} is trained using a cross-entropy loss to model the distribution of generated tokens $x_{\theta}$
and to form the gradient to update $\theta$. \yzc{The teacher $\phi$ is frozen during training.}
}
\label{fig:pipeline}
\vspace{-0.3cm}
\end{figure*}
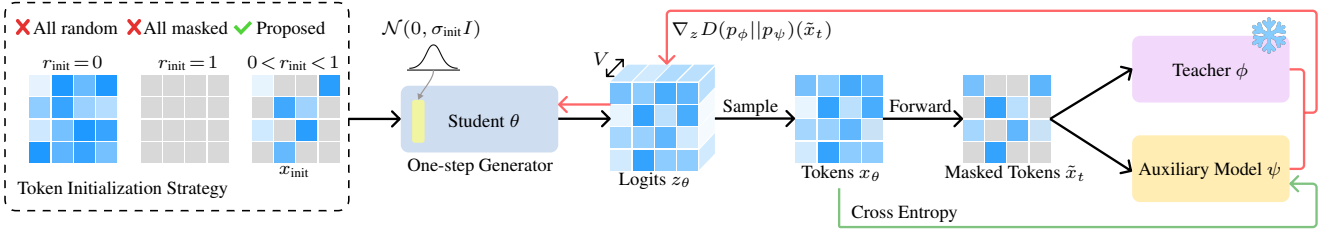

\section{Method}
\label{sec:method}

We introduce \textbf{Di}stilling  $\mathtt{[M]}$DM into \textbf{O}ne-step generator (\method{}), a novel method for distilling a multi-step MDM teacher model $\phi$ into a one-step generator $\theta$. 
An overview of \method{} is illustrated in \cref{fig:pipeline}.  
The first step is to define the initial distillation objective, which aims to align the generation distribution of the one-step generator and the multi-step teacher in \cref{sec:token_matching}.
In \cref{sec:loss_gradient}, we introduce an approximation strategy that formulates an effective gradient {to update} the generator.
In \cref{sec:init_code}, we propose initialization strategies to address the low-entropy issue of the initial masked state in MDMs.
By combining these components, our method enables efficient and high-fidelity discrete data generation, bridging the gap between multi-step MDMs and one-step inference models.

\subsection{\yzc{One-step On-policy Distillation}}
\label{sec:token_matching}

Our goal is to train the student model to transform an initial random input distribution, $p_{\text{init}}$, into outputs that closely resemble the teacher’s multi-step generation distribution, $p_\phi$ (note that we discuss the specific design of $p_{\text{init}}$ in Section \cref{sec:init_code}.).
Taking inspiration from recent works on on-policy distillation \cite{agarwal2024policy,ross2011reduction,ho2016model,balakrishna2020policy,arora2022exposure,schmidt2019generalization,hinton2015distilling}, we propose to first sample intermediate state $\tilde{x}_t$ derived from the student’s own predictions and then match the conditional predicted distributions of teacher $p_\phi(x_0 | \tilde{x}_t)$ and student $p_\theta(x_0 | \tilde{x}_t)$.
We denote this procedure as ${D}(p_\phi||p_\theta)(\tilde{x}_t):=D\big(p_\phi({x_0}|\tilde{x}_t) \| p_\theta({x_0}|\tilde{x}_t)\big)$, where $D$ is the divergence. 
By enforcing this match for all possible intermediate states $\tilde{x}_t$, we effectively ensure that the overall generation distribution of the student model aligns with that of the teacher.
The overall distillation objective can be formulated as follows:
 \begin{equation}
 \label{eq:on_policy}
 \begin{aligned}
     \mathcal{L}_\text{\method{}}(\theta) := \mathbb{E}_{x_{\text{init}}, t} \left[ w(t)\left(\mathbb{E}_{q_{t|0}} [{D}(p_\phi||p_\theta)(\tilde{x}_t)] \right)\right],
 \end{aligned}
 \end{equation}
where $t\sim \mathcal{U}[0,1]$ and $w(t)$ is a weighting function.
Since the generator $\theta$ can now only take $x_{\text{init}}$ as input and directly produces $x_0$, we apply the forward diffusion process to get $\tilde{x}_t \sim q_{t|0}(\tilde{x}_t | x_\theta(x_\text{init}))$ as the pseudo-intermediate state \yzc{from its output $x_\theta(x_\text{init})$}.
This divergence is calculated across all possible $x_\text{init}$ and $t$ to cover every possible intermediate state.

\vspace{0.2cm}
\noindent\textbf{Token-level Distribution Matching.} 
The conditional divergence in \cref{eq:on_policy} can be further decomposed and \yzc{formulated} as the averaged divergence of each masked tokens given $\tilde{x}_t$, defined as:
\begin{equation}\label{eq:mdm_div}
  D\big(p_\phi \| p_\theta\big)(\tilde{x}_t) := \frac{1}{L_M} \sum_{\substack{i=1 \\ {\tilde{x}_t^i} = \mask{}}}^L 
 D\big(p_\phi({x_0^i}|\tilde{x}_t) \| p_\theta({x_0^i}|\tilde{x}_t)\big),
\end{equation}
where $L_M$ is the number of the mask tokens in $\tilde{x}_t$.
Unlike {in} continuous diffusion \yzc{model} distillation \cite{yin2024one,luo2023diff,nguyen2024swiftbrush}, the divergence is computed over the fully generated latents of the models, our proposed \method{} instead calculates divergence at the \emph{token level} \yzc{with the decomposition in \cref{eq:mdm_div}}.

\subsection{Approximation of Loss Gradient}
\label{sec:loss_gradient}

Directly optimizing \cref{eq:on_policy} is not feasible because the exact form of $p_\theta(x_0^i|\tilde{x}_t)$ is unknown. 
Similar to prior work \cite{poole2022dreamfusion, huang2024flow, zhou2024score}, we seek to approximate the following gradient of training objective (see \cref{supp:loss_grad}):
\begin{equation}\label{eq:D-grad}
\footnotesize
\begin{aligned}
\nabla&_\theta\mathcal{L}_\text{\method{}} \!=\!
    \mathbb{E}_{x_{\text{init}}, t}\!\!\left[w(t)\!\!\left(\!\mathbb{E}_{q_{t|0}}\!\!\left[\colororange{\nabla_{{z_\theta}} {{D}(p_\phi||p_\theta)(\tilde{x}_t)}} \,\colorblue{\frac{\text{d} {z_\theta}(\tilde{x}_t)}{\text{d} \theta}}\right] \!\right) \!\!\right]\!,
\end{aligned}
\end{equation}
where the logits ${z_\theta}$ is the direct output of {student} model.
In this way, we can view the loss gradient as the product of two terms: the gradient of the divergence with respect to model logits (in orange), and the Jacobian of the model output with respect to {student} parameters $\theta$ (in blue).
By approximating these two terms separately, we can address the intractability of directly optimizing the original objective:
\begin{enumerate}
    \item \colorbox{orange!15}{Approximation of $p_\theta(x_0^i|\tilde{x}_t)$:}
    Given that the highlighted divergence gradient term in \cref{eq:D-grad} can be viewed as a functional of the teacher output $p_\phi(x_0^i|\tilde{x}_t)$ and {the} unknown student output $p_\theta(x_0^i|\tilde{x}_t)$, 
    we adopt the strategy from \cite{wang2024prolificdreamer,luo2023diff,yin2024one} and introduce an \fakemodel{} to approximate the unknown $p_\theta(x_0^i|\tilde{x}_t)$. 
    The \fakemodel{} $\psi$ is trained as a $x_0$-predictor using data generated by the one-step generator and is optimized via the MDM training loss in \cref{eq:MDM_loss} to have $p_\psi(x_0^i|\tilde{x}_t) \approx p_\theta(x_0^i|\tilde{x}_t)$.
    
    \item \colorbox{blue!8}{Approximation of $z_\theta(\tilde{x}_t)$:}
    The approximation of the model's Jacobian term must satisfy a critical requirement: it needs to provide gradient information to update the model parameters $\theta$.
    Given this constraint, we propose approximating the intractable implicit term $z_\theta(\tilde{x}_t)$ by substituting it with the one-step output $z_\theta(x_\text{init})$, under the assumption of {consistency property} \cite{song2023consistency} \yzc{(see \cref{fig:consistency_assump} for illustration)}.
    
\end{enumerate}
\noindent These approximations lead to the following gradient of loss:
\begin{equation}\label{eq:D-grad_approx}
\footnotesize
\begin{aligned}
\nabla&_\theta\mathcal{L}_\text{\method{}} 
    \!\approx\! 
    \mathbb{E}_{x_{\text{init}}, t}\!\!\left[w(t)\! \!\left(\!\mathbb{E}_{q_{t|0}}\!\!\left[\colororange{\nabla_{z_\psi} {D}(p_\phi||p_\psi)(\tilde{x}_t)}\,\colorblue{\frac{\text{d} {z_\theta}(x_\text{init})}{\text{d} \theta}}\right]\!\right)\!\!\right]\!.
\end{aligned}
\end{equation}
To effectively transfer prior knowledge from teacher model, we initialize both the generator $\theta$ and \fakemodel{} $\psi$ with the pre-trained weights of the teacher model $\phi$. 

\vspace{0.2cm}
\noindent\textbf{Generalized Jeffrey Divergence.}
\yzc{Unlike similar works on continuous diffusion distillation~\cite{luo2023diff,yin2024one,nguyen2024swiftbrush} {that} rely on the Reverse KL (RKL) divergence, the proposed \method{} is not limited to this choice. Instead, it can leverage alternative token-level divergence measures, as we do not impose restrictions on the form of divergence in \cref{eq:D-grad_approx}}.
This is an especially interesting feature as the  RKL is known to exhibit unwanted mode-seeking behaviors~\cite{dao2025swiftbrush}. 

In particular, we propose to explore the use of Generalized Jeffrey Divergence \cite{jeffreys1946invariant, sessa2024bond}, which is defined as a linear combination of the Forward KL (FKL) and RKL divergences and investigate its effect beyond mode-seeking (see \cref{supp:mode_seeking}): 
\begin{equation}\label{eq:jeffrey}
\begin{aligned}
D_{\text{Jeffrey}}^\beta = (1-\beta) {D}_{FKL} + \beta {D}_{RKL},
\end{aligned}
\end{equation}
with $\beta$ the hyperparameter. 
The explicit gradient of these divergences is provided in \cref{supp:div_grad}.

\subsection{Token Initialization Strategy}
\label{sec:init_code}
Unlike continuous diffusion models, where the initial distribution is typically a standard Gaussian, the initial state $x_1$ of MDM consists of a \textit{sequence} of deterministic identical $\mask{}$ tokens \cite{austin2021structured,hayakawa2024distillation}.
This fixed initialization leads to mode collapse for one-step generation, as the generator can only output fixed logits $\hat{z}_\theta$ with limited sample diversity.
We therefore explore three possible strategies for the token sequence initialization, including the fixed one, as follows (see left panel of \cref{fig:pipeline}):
\begin{enumerate}
    \item \textbf{All masked tokens:} following the standard MDM initialization \cite{shi2025simplified,sahoo2024simple}, we initialize the sequence with only $\mask{}$ tokens.
    \item \textbf{All random tokens:} 
    aiming for maximizing entropy, we choose image tokens with random values to fill the initial token sequence. 
    \item \textbf{Hybrid strategy}: combining the two previous strategies by fixing an initial mask ratio $r_{\text{init}}$ and replacing the remaining $1 {-} r_{\text{init}}$ of the $\mask{}$ tokens with random {image} tokens.
\end{enumerate}

\noindent
In our work, we adopt the \textbf{hybrid strategy}, guided by two key hypotheses and empirically supported by our ablation analysis in~\cref{fig:DDMD_Ablation}: (1) To prevent mode collapse, the initialization of one-step generator should include a certain level of \emph{randomness}; (2) as the one-step generator is inherited from the teacher, whose objective is to predict all tokens from \emph{partially masked inputs}, the initialization must preserve a similar {pattern}, i.e., include \emph{some mask tokens} to avoid input distribution mismatch. 

Moreover, inspired by prior {auto-regressive} distillation work~\cite{liu2024distilled}, which proposed to replace all initial token embeddings with random Gaussian noise, we therefore add Gaussian perturbation upon our current scheme to further increase the randomness from a discrete space to a real one. Specifically, after randomly selecting $ r_{\text{init}} $ $\mask{}$ tokens and $ 1 {-} r_{\text{init}} $ random image tokens, we perturb all token embeddings $\mathbf{e}$ with Gaussian noise $\epsilon$ at a fixed noise level $\sigma_{\text{init}}$ using a variance-preserving scheme \cite{ho2020denoising,song2020score} $\hat{\mathbf{e}} = \sqrt{1 - \sigma_{\text{init}}^2}\,\mathbf{e} + \sigma_{\text{init}}\,\epsilon$. 
as illustrated in Figure~\ref{fig:pipeline}. 

\vspace{0.2cm}
\noindent\textbf{Overview of Algorithm.}
We now summarize our method in \cref{alg:distill} where the generator and \fakemodel{} are updated iteratively. In each iteration, the generator produces tokens $x_\theta$ from initial state $x_{\text{init}}$ in one step and is optimized with the loss gradient in \cref{eq:D-grad_approx}, while the \fakemodel{} is trained on these generated tokens as targets.

\begin{algorithm}[t]
    \small
    \caption{\method{} Distillation}
    \label{alg:distill}
    \begin{algorithmic}[1]
    \Require Pre-trained teacher model ${\phi}$, condition dataset $\mathcal{D}$
    \State{$\theta \leftarrow \text{copyWeights}(\phi),$
    $\psi \leftarrow \text{copyWeights}(\phi)$ \textcolor[rgb]{0.40,0.40,0.40}{{$\,\,\,$// intialize}}}
    \Repeat
        \State{\textcolor[rgb]{0,0.5,0}{\textit{// Generate tokens $x_0$}}}
        \State{Sample $x_{\text{init}}\sim p_{\text{init}}$, $c\sim \mathcal{D}$ \textcolor[rgb]{0.40,0.40,0.40}{$\,\,$// with strategy in \cref{sec:init_code}}}
        \State{Get generator logits $z_\theta(x_{\text{init}},c) \in \mathbb{R}^{B\times h\times w\times V}$ }
        \State{$x_\theta\!\in \!\mathbb{R}^{B\times h\times w} \xleftarrow[]{\text{sample}} p_\theta(x_0|x_{\text{init}})\!=\!\text{softmax}(z_\theta(x_{\text{init}},c))$ 
        }
        \State{\textcolor[rgb]{0,0.5,0}{\textit{// Update generator ${\theta}$}}}
        \State{Sample $t\!\sim\!\mathcal{U}[0,1]$, $\tilde{x}_t \!\sim\! q_{t|0}(\tilde{x}_t | x_\theta(x_\text{init},c))$ \textcolor[rgb]{0.40,0.40,0.40}{$\,\,\,$// Forward}}
        \State{Calculate $p_\phi({x_0}|\tilde{x}_t,c)$ and $p_\psi({x_0}|\tilde{x}_t,c)$}
        \State{Update ${\theta}$ with the loss gradient $\nabla_\theta\mathcal{L}_\text{\method{}}$ (\cref{eq:D-grad_approx})}
        \State{\textcolor[rgb]{0,0.5,0}{\textit{// Update \fakemodel{} ${\psi}$}}}
        \State{Sample $t'\!\sim\!\mathcal{U}[0,1]$, $\tilde{x}_{t'} \!\sim\! q_{t|0}(\tilde{x}_{t'} | x_\theta(x_\text{init},c))$ }
        \State{Update ${\psi}$ with cross entropy loss (\cref{eq:MDM_loss})}
    \Until{\textit{convergence}}
    \State{\textbf{Return} one-step generator ${\theta}$}
    \end{algorithmic}
\end{algorithm}

\section{Experiments}
\label{sec:experiments}

\noindent\textbf{Teacher Models.}
\xwc{We conduct extensive distillation experiments on both class-conditional image generation and text-to-image generation. For class-conditional generation, we adopt the best-performing and publicly available class-conditional MDM, MaskGit \cite{besnier2023pytorch}, as the teacher model. For text-to-image generation, we employ the recent Meissonic \cite{bai2024meissonic} as the teacher and use the LAION-Aesthetics-6+ prompt dataset \cite{cherti2023reproducible} for prompting during the distillation.}

\vspace{0.1cm}
\noindent\textbf{Evaluation Metrics.}
The metrics we use for comparing ImageNet results are Fr\'echet Inception Distance (FID) \cite{heusel2017gans} and Inception Score (IS) \cite{salimans2016improved}.
\yzc{We also use precision (Prec.) and recall (Rec.) \cite{kynkaanniemi2019improved}, density (Den.) and
coverage (Cov.) \cite{naeem2020reliable} to further evaluate the fidelity and diversity of generated images.}
For text-to-image generation, we {follow~\cite{bai2024meissonic}} and measure the HPSv2 \cite{wu2023human} and Geneval \cite{ghosh2023geneval} score. {See \cref{supp:more_exps} for more metrics and results of both tasks.}
In our experiments on ImageNet 256$\times$256, we calculate the FID using 5k generated images for ablations and 50k generated for benchmarking, comparing them against 50k images from the ImageNet validation set with Clean-FID \cite{parmar2022aliased}.

\vspace{0.1cm}
\noindent\textbf{Experiment Setup.}
Both teacher models {adopt} Classifier-Free Guidance (CFG) \cite{ho2022classifier}.
For the MaskGit teacher, we maintain a CFG scale of 2 during distillation, while for the Meissonic teacher, we use a CFG scale of 4. 
Our generator is always trained with conditioning, eliminating the need for CFG during inference, further improving the sampling efficiency.
All of the ablation experiments on MaskGit are trained with a batch size of 64 and 30000 iterations.
We evaluate the final checkpoints with different temperature to get the best metric results.
The learning rates for MaskGit experiments and Meissonic experiments are $1\!\times\!10^{-5}$ and $1\!\times\!10^{-6}$, respectively.
See \cref{supp:expr_setup} for more details.

\subsection{Class-conditional Image Generation}

In \cref{table:imagenet}, we present a quantitative performance comparison of our \mname{} with various acceleration methods, including the high-order sampler $\theta$-trapezoidal \cite{ren2025fast}, other distillation techniques, and the teacher model. 
Compared to the high-order sampler, which requires 32 steps to achieve an FID of 7.1, our method significantly reduces the number of steps to just 1 while maintaining competitive performance, with an FID of 6.91 and an IS of 214.0.
When comparing to distillation methods, our approach shows comparable results to di4c \cite{hayakawa2024distillation}, which uses 4 steps and achieves an FID of 6.79 and an IS of 209.2. Our method yields an FID of 6.91 with a higher IS of 214.0 in one step.
Additionally, when compared to the teacher model (MaskGit \cite{besnier2023pytorch}) with varying steps, our method performs very closely to the teacher's performance of 6.60 FID at 16 steps, while achieving this with just a single step. 
These comparisons highlights the effectiveness of our method in matching the performance of multi-step teacher models while dramatically reducing the computational cost. 
For further details, refer to \cref{supp:expr_setup}.

\begin{figure*}[t!]
\vspace{-0.6cm}
\centering
\begin{overpic}[width=1.\linewidth]{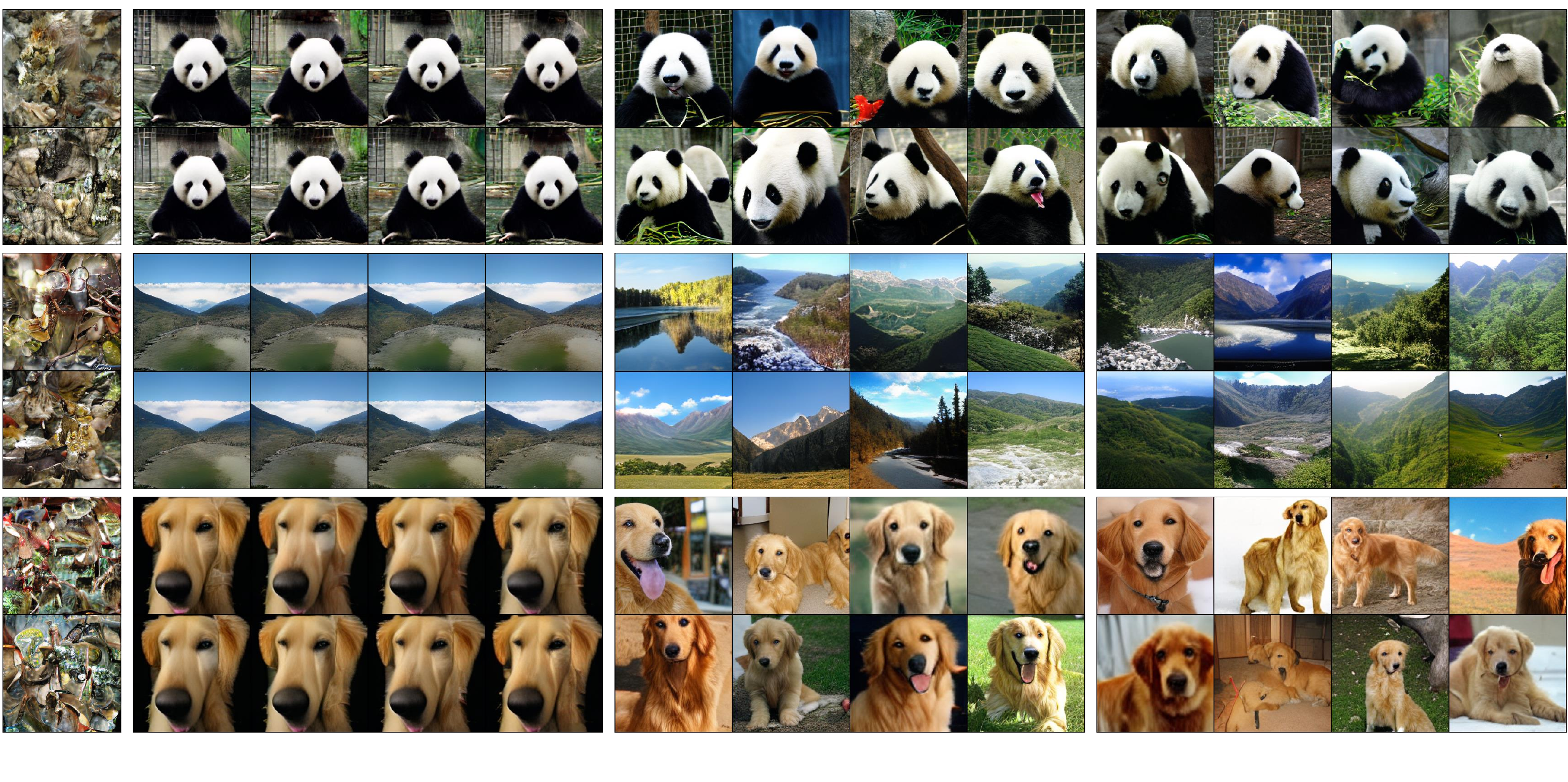}
\put(1.,1.4){\color{black}{\footnotesize $r_{\text{init}}=0$}}
\put(13.5,1.4){\color{black}{\footnotesize \method{} ($r_{\text{init}}=1$) one step }}
\put(44.8,1.4){\color{black}{\footnotesize{\method{} ($r_{\text{init}}=0.6$) one step }}}
\put(79.5,1.4){\color{black}{\footnotesize{Teacher 16 steps}}}
\end{overpic}
\vspace{-0.6cm}
\caption{\textbf{Visual results of ImageNet.} 
One-step generated images from the generator trained with different $r_{\text{init}}$ in comparison with teacher generation with 16 sampling steps.
The class labels of the samples from top to bottom are \texttt{388}, \texttt{979} and \texttt{207} respectively.
}
\label{fig:mode_collapse}
\vspace{-0.2cm}
\end{figure*}

\begin{figure*}[ht]
    \centering
    \begin{subfigure}[b]{0.335\textwidth}
        \centering
         \includegraphics[width=\textwidth]{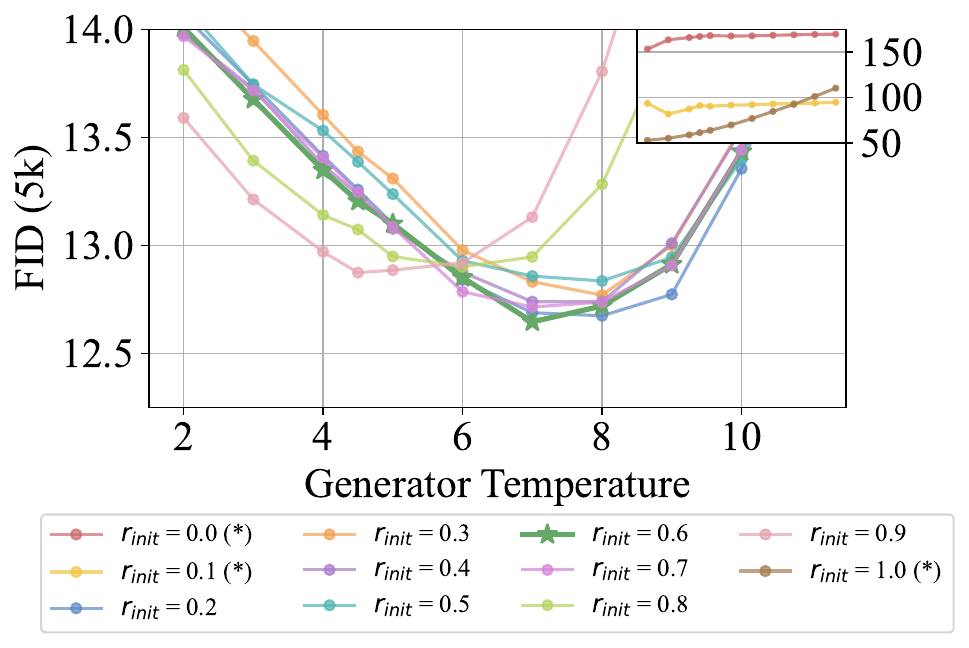}
        \subcaption{{$r_{\text{init}}$}: Initial Mask Ratio}
        \label{subfig:DDMD_Ablation_ratio}
    \end{subfigure}
    \hfill
    \begin{subfigure}[b]{0.32\textwidth}
        \centering
         \includegraphics[width=\textwidth]{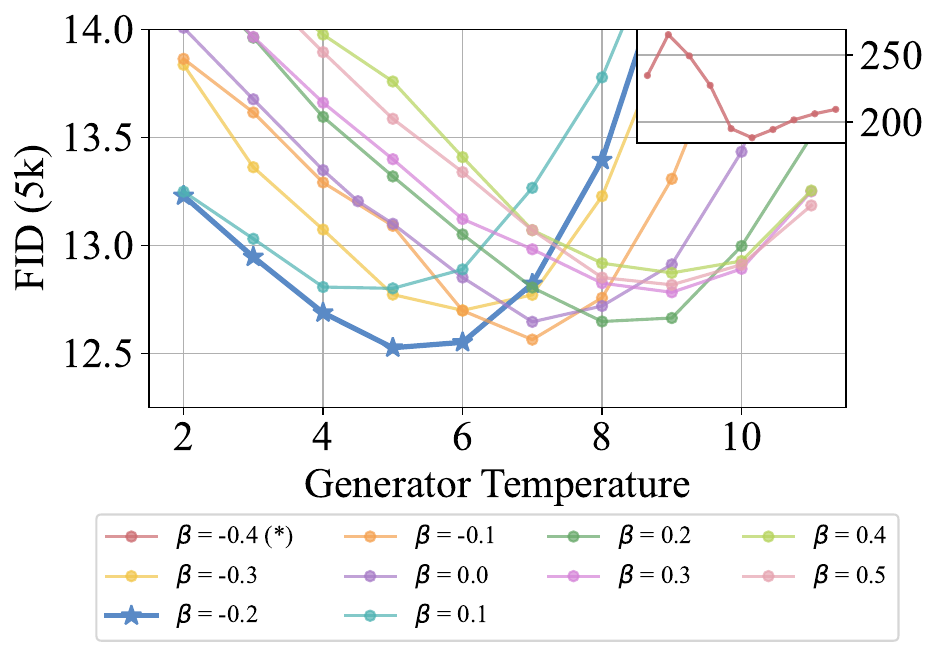}
        \subcaption{\textbf{$\beta$}: Jeffrey Coefficient}
        \label{subfig:DDMD_Ablation_beta}
    \end{subfigure}
    \hfill
    \begin{subfigure}[b]{0.335\textwidth}
        \centering
         \includegraphics[width=\textwidth]{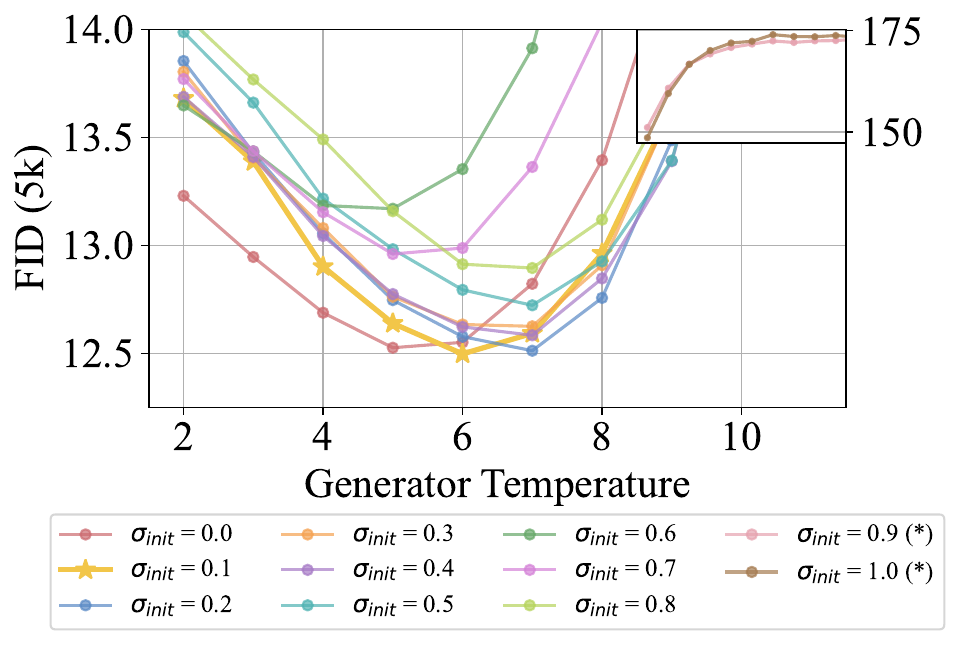}
        \subcaption{$\sigma_{\text{init}}$: Perturbation Strength}
        \label{subfig:DDMD_Ablation_noise_pert}
    \end{subfigure}
    \caption{Ablation studies on ImageNet using FID as the evaluation metric. $^*$ means {the training is collapsed and falls outside the comparable range with other results, we show these in the sub-figures at the right upper corner with the same x-axis range.}
    }
    \label{fig:DDMD_Ablation}
    \vspace{-0.3cm}
\end{figure*}

\begin{table}[!t]
\centering
\caption{
Quantitative results on \yzc{class-conditional ImageNet-256}. $^*$ denotes numbers estimated from the original {plot}.
}
\vspace{-0.2cm}
\resizebox{0.99\linewidth}{!}{
\setlength{\tabcolsep}{3pt}
\begin{tabular}{llccccccc}
\toprule
ImageNet & Method & Step ($\downarrow$)  & FID ($\downarrow$) & IS ($\uparrow$)  & Prec. ($\uparrow$)  & Rec.  ($\uparrow$)  & Den.  ($\uparrow$)  & Cov.  ($\uparrow$)   \\
\midrule 
\multirow{4}{*}{\shortstack[l]{{Teacher}}} 
& MaskGit \cite{besnier2023pytorch} & 16 & 6.60 & 224.07 & 0.831 &  0.402 &  1.246 &  0.977 \\
& MaskGit \cite{besnier2023pytorch} & 8 & 6.66 & 221.57 & 0.827 & 0.397 & 1.233 & 0.974 \\
& MaskGit \cite{besnier2023pytorch} & 4 & 10.73 & 192.29 & 0.748 &  0.313 &  1.011 &  0.920 \\
& MaskGit \cite{besnier2023pytorch} & 2 & 91.35 & 13.37 & 0.178 &  0.164 &  0.091 &  0.122 \\
\midrule 
\multirow{2}{*}{\shortstack[l]{{Sampler}}} 
& $\theta$-trapezoidal \cite{ren2025fast}$^*$ & 64 & 6.7 & - & - & - & - & - \\
& $\theta$-trapezoidal \cite{ren2025fast}$^*$ & 32 & 7.1 & - & - & - & - & -  \\
\midrule 
\multirow{4}{*}{\shortstack[l]{Distillation}} 
& di4c \cite{hayakawa2024distillation} & 4 & 
6.79 & 209.2  & - & - & - & - \\
& di4c-d \cite{hayakawa2024distillation} & 4 & 6.57 &  213.6 & - & - & - & - \\
& \cellcolor{gray!20} \mname{} & \cellcolor{gray!20} \textbf{1} & \cellcolor{gray!20} 6.91 & \cellcolor{gray!20} 214.0 & \cellcolor{gray!20} 0.828 & \cellcolor{gray!20} 0.377 & \cellcolor{gray!20} 1.255 & \cellcolor{gray!20} 0.967 \\ 
\bottomrule
\end{tabular}}
\label{table:imagenet}
\vspace{-0.2cm}
\end{table}

\subsection{Ablations}

{In this section, we present ablation studies on key hyperparameters of our method to validate our design choices: (1) the initial mask ratio \(r_{\text{init}}\); (2) the Jeffrey Coefficient \(\beta\); and (3) the Gaussian perturbation Strength \(\sigma_{\text{init}}\):}

\vspace{0.2cm}
\noindent\textbf{Initial Mask Ratio  $r_{\text{init}}$.}
We begin our ablation study by analyzing the impact of the initial mask ratio $r_{\text{init}}$, which is a crucial factor in our algorithm.
As shown in \cref{subfig:DDMD_Ablation_ratio}, our results confirm the hypothesis from \cref{sec:init_code} that both extreme values, $r_{\text{init}} = 0$ and $r_{\text{init}} = 1$ fail to work. 
Specifically, setting $r_{\text{init}} = 1$ results in mode collapse, while very low values ($r_{\text{init}} \approx 0$) cause unstable training, as evidenced by the broken curves \yzc{in the subfigures of} \cref{subfig:DDMD_Ablation_ratio}. 
The optimal choice appears to be $r_{\text{init}} = 0.6$, which achieves the lowest FID. See \cref{fig:mode_collapse} for visual demonstrations.

\vspace{0.2cm}
\noindent\textbf{Jeffrey Coefficient $\beta$.}
Building on this, we conduct an ablation on the generalized Jeffrey divergence coefficient $\beta$ (\cref{subfig:DDMD_Ablation_beta}). Our experiments indicate that reducing $\beta$ generally improves FID, aligning with observations from prior work \cite{xu2025one}. Surprisingly, our method remains effective even for negative values of $\beta$, with the best performance observed at $\beta = -0.2$.

\vspace{0.2cm}
\noindent\textbf{Gaussian Perturbation $\sigma_{\text{init}}$.}
Finally, we examine the effect of Gaussian perturbation on token embeddings (\cref{subfig:DDMD_Ablation_noise_pert}). Our results demonstrate that introducing perturbations provides additional improvements in FID, further refining the quality of the generated samples.
For completeness, the corresponding results with the IS metric are provided in \cref{fig:DDMD_Ablation_IS} in the Appendix.

\subsection{Text-to-image Generation}

\begin{figure*}
    \vspace{-0.5cm}
    \centering
    \includegraphics[width=1.0\linewidth]{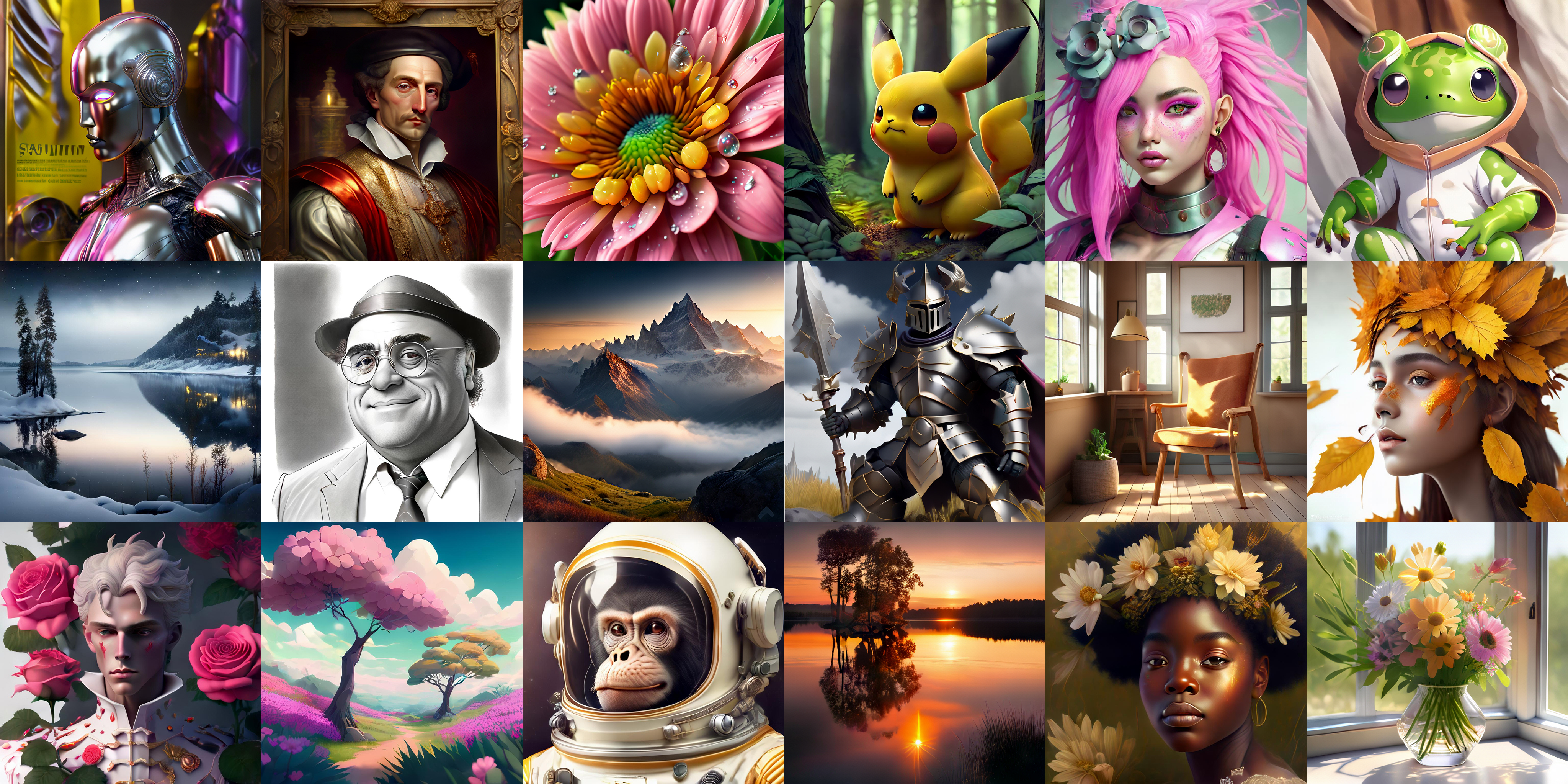}
    \vspace{-0.5cm}
    \caption{Qualitative our method distilled from teacher Meissonic. \yzc{Corresponding prompts can be found in \cref{supp:Prompts}}}
    \label{fig:qualitative_meissonic}
    \vspace{-0.2cm}
\end{figure*}

{We validate \mname{} for text-to-image generation by evaluating our distilled one-step generator on the HPSv2~\cite{wu2023human} and GenEval~\cite{ghosh2023geneval} benchmarks.}

{In~\cref{tab:benchmark-hps-v2}, we show the HPSv2~\cite{wu2023human} benchmark evaluating human preference for T2I models. Distilled from Meissonic~\cite{bai2024meissonic}, the only open-sourced text-to-image MDM, we compare our one-step generator with not only the teacher model, but also \yzc{recent} diffusion models~\cite{rombach2022high,dhariwal2021diffusion,podell2024sdxl,deepfloyddeepfloyd} and their distilled one-step generators~\cite{liu2023instaflow,sauer2024adversarial,dao2025swiftbrush}.
The results clearly show that our one-step generator achieves a competitive performance with teacher models generated using 16 to 32 steps, whereas teacher's performance deteriorates rapidly when fewer generation steps: a 48-step generation yields $28.83$, whereas the 4 steps' drop to $24.66$). Our one-step generator also outperforms other one-step continuous diffusion-based ones; however, direct comparisons are limited since they are distilled from different teachers. 

{To better evaluate the semantic expressiveness of our one-step generator, \cref{tab:genEvalSotaTable} presents a comparison with both diffusion models~\cite{rombach2022high,dhariwal2021diffusion,podell2024sdxl,ramesh2022hierarchical} and our teacher, Meissonic~\cite{bai2024meissonic}. The experimental results corroborate our previous observations: our one-step generator competes with teacher models' performance using 16 to 32 steps, while the teacher's performance degrades rapidly with reduced steps.}

\begin{table}[t]
  \caption{HPS v2.0 benchmark. Scores are collected from \url{https://github.com/tgxs002/HPSv2}. We highlight the \textbf{best}.}
  \label{tab:benchmark-hps-v2}
  \centering
  \resizebox{1.0\linewidth}{!}{\setlength{\tabcolsep}{2pt}
  \begin{tabular}{lcccccc}
    \toprule
    \multicolumn{7}{c}{\bf{HPS v2.0}}                   \\
    \midrule
    \bf{Model} & \bf{Step} &\bf{Anim.} & \bf{Concept-art} & \bf{Painting} & \bf{Photo} & \bf{Averaged}\\
    \midrule
    Latent Diffusion~\cite{rombach2022high} & 25 & $25.73$ & $25.15$ & $25.25$ & $26.97$ & $25.78$ \\
    DALL·E 2~\citep{ramesh2022hierarchical} & - & $27.34$ & $26.54$ & $26.68$ & $27.24$ & $26.95$ \\
    Stable Diffusion v1.4~\cite{rombach2022high} & 50   & $27.26$ & $26.61$ & $26.66$ & $27.27$ & $26.95$ \\
    Stable Diffusion v2.0~\cite{rombach2022high} & 50  & $27.48$ & $26.89$ & $26.86$ & $27.46$ & $27.17$ \\
    DeepFloyd-XL~\citep{deepfloyddeepfloyd}  & 25  & $27.64$ & $26.83$ & $26.86$ & $27.75$ & $27.27$ \\
    SDXL Base 1.0~\cite{podell2024sdxl}  & 50  & $28.88$ & $27.88$ & $27.92$ & $28.31$ & $28.25$ \\
    SDXL Refiner 1.0~\citep{podell2024sdxl} & 50   & $28.93$ & $27.89$ & $27.90$ & $28.38$ & $28.27$ \\
    \midrule
    InstaFlow \cite{liu2023instaflow} & 1 & 25.98 & 25.79 & 25.93 & 26.32 &  26.01 \\
    SD Turbo \cite{sauer2024adversarial} & 1 & 27.98 & 27.59 & 27.16 & 27.19  & 27.48 \\
    SwiftBrush v2 \cite{dao2025swiftbrush} & 1 & 27.25 & 27.62 & 26.86 & 26.77 & 27.15 \\
    \midrule
    \multirow{6}{*}{Meissonic \cite{bai2024meissonic}}  & 48 & $\textbf{29.57}$ & $\textbf{28.58}$ & $\textbf{28.72}$ & $\textbf{28.45}$ & $\textbf{28.83}$ \\ %
     & 32  & $29.18$ & $28.32$ & $28.28$ & $27.96$ & $28.44$ \\ 
     & 16  & $28.61$ & $27.82$ & $27.84$ & $27.32$ & $27.90$ \\ 
     & 8  & $25.62$  & $26.49$ & $26.67$ & $27.07$ & $26.46$   \\ 
     & 4  & $25.01$ & $24.95$ & $24.87$ & $23.80$ & $24.66$ \\ 
     & 2  & $23.06$ & $23.28$ & $23.22$ & $22.38$ & $22.98$ \\ 
    \midrule
    \rowcolor{gray!20}
    {\mname{}} & 1  & $28.64$ & $27.91$ & $27.99$ & $27.92$ & $28.11$ \\ 
    \bottomrule
  \end{tabular}
  }
\vspace{-0.2cm}
\end{table}

\begin{table}[t]
\centering
\caption{GenEval benchmark. We highlight the \textbf{best} result.}
\label{tab:genEvalSotaTable}
\resizebox{1.\linewidth}{!}{\setlength{\tabcolsep}{2pt}
\begin{tabular}{lcccccccc}
    \toprule
    \multicolumn{9}{c}{\bf{GenEval}}                   \\
    \midrule
    \multirow{2}{*}{\bf Model}  & \multirow{2}{*}{\bf Step} & \multirow{2}{*}{\bf Overall} & \multicolumn{2}{c}{\bf Objects} & \multirow{2}{*}{\bf Counting} & \multirow{2}{*}{\bf Colors} & \multirow{2}{*}{\bf Position} & \multirow{2}{*}{\bf \shortstack[l]{ {Color} \\ {Attribution} }} \\
    \cmidrule(lr){4-5}
     &  & & {\bf Single} & {\bf Two} &  &  &  &  \\
    \midrule
    SD v1.5~\cite{rombach2022high} & 50 & 0.43 & 0.97 & 0.38 & 0.35 & 0.76 & 0.04 & 0.06 \\
    SD v2.1~\cite{rombach2022high} & 50 & 0.50 & 0.98 & 0.51 & 0.44 & 0.85 & 0.07 & 0.17 \\
    DALL-E2~\cite{ramesh2022hierarchical}& - & 0.52 & 0.94 & 0.66 & \textbf{0.49} & 0.77 & 0.10 & 0.19 \\
    SDXL~\cite{podell2024sdxl}  & 50 & \textbf{0.55} & 0.98 & \textbf{0.74} & 0.39 & 0.85 & \textbf{0.15} & \textbf{0.23} \\
    \midrule
    \multirow{6}{*}{Meissonic \cite{bai2024meissonic}}  & 48 & 0.54 &\textbf{0.99} & 0.66 & 0.42 & \textbf{0.86} & 0.10 & 0.22 \\ %
     & 32 & 0.46 & 0.92 & 0.53 & 0.33 & 0.80 & 0.08 & 0.13 \\
     & 16 & 0.37 & 0.82 & 0.39 & 0.20 & 0.70 & 0.05 & 0.08 \\
     & 8 & 0.20 & 0.58 & 0.12 & 0.05 & 0.40 & 0.02 & 0.04 \\
     & 4 & 0.09 & 0.31 & 0.02 & 0.01 & 0.18 & 0.01 & 0.01 \\
     & 2 & 0.03 & 0.14 & 0.01 & 0.00 & 0.05 & 0.00 & 0.00 \\
    \rowcolor{gray!20}
    \midrule
    {\mname{}} & 1 & 0.43 & 0.91 & 0.53 & 0.22 & 0.75 & 0.07 & 0.11 \\
    \bottomrule
\end{tabular}
}
\vspace{-0.2cm}
\end{table}

\vspace{0.2cm}
\noindent\textbf{Limitations and Future Works.}
While our data-free distillation achieves teacher-level performances, its application has so far been limited to our current model scope. In the future, we aim to extend our approach to stronger MDM teachers, particularly for image and text generation. 
While our method is data-free, it would be beneficial to also introduce real data to boost the one-step student to outperform the teacher.
Lastly, while we leverage generalized Jeffreys divergence to avoid mode seeking behaviors, we intend to explore more general $f$-divergences to increase our method's flexibility and effectiveness.

\section{Conclusion}
\label{sec:conclusion}
In this work, we proposed \method{}, a novel method that leverages token-level distribution matching to distill the inference process of MDMs into one-step.
Specifically, inspired by the concept of on-policy distillation, we match the token-level distribution conditioned on a pseudo intermediate state obtained from the student's one-step generation. 
We also conducted extensive experiments on the choice of initialization strategy and distillation objective, to increase the robustness of \mname{}.
Our experimental results demonstrate that our distilled model generates images with comparable quality to the teacher models, while requiring only 1 sampling step during inference.
This work demonstrates the power of distribution matching methods on distillation for MDM, contributes to the growing community of research exploring efficient generation of discrete data.

\clearpage
\newpage

\section*{Acknowledgements} 
\xwc{This work was supported by ANR-22-CE23-0007, ANR-22-CE39-0016, Hi!Paris grant and fellowship, DATAIA Convergence Institute as part of the “Programme d’Investissement d’Avenir” (ANR-17-CONV-0003) operated by Ecole Polytechnique, IP Paris, and was granted access to the IDRIS High-Performance Computing (HPC) resources under the allocation 2024-AD011014300R1 and 2025-AD011015894} made by GENCI and {mesoGIP of IP Paris}. 
We also sincerely thank {Nacereddine Laddaoui for the help with infrastructure}, Haoge Deng and Yao Teng for their insightful discussions that contributed to this work. We are also grateful to Nicolas Dufour, Robin Courant, and Lucas Degeorge for their meticulous proofreading. 

\section*{Broader Impacts}
Our work focuses on distilling the multi-step generation process of MDMs into one step, significantly reducing inference time and computational costs, {therefore lowering the carbon footprint during the inference}. This advancement has the potential to make high-quality generative models more accessible, facilitating applications in creative industries, content generation, and real-time systems.
However, as with many generative modeling techniques, our method inherits biases from the teacher models. This could potentially lead to ethical concerns, including the generation of misleading or harmful content. 
Additionally, by enabling faster and more efficient content generation, our approach could lower the barrier to misuse, such as the creation of deepfakes or other deceptive media.

{\small
\bibliographystyle{ieeenat_fullname}
\bibliography{main}
}

\clearpage
\onecolumn  
\setcounter{section}{0}
\renewcommand{\thesection}{\Alph{section}}
\appendix
\setcounter{page}{1}

\begin{center}
    \Large
    \textbf{\thetitle}\\
    \vspace{0.5em}Supplementary Material \\
    \vspace{1.0em}
\end{center}

The supplementary material is organized as follows:
\begin{itemize}
    \item \cref{supp:derivation}: Relevant derivations of the approximated divergence gradient.
    \item \cref{supp:related}: Discussion of additional related works.
    \item \cref{supp:expr_setup}: Detailed experiment setup.
    \item \cref{supp:more_exps}: Additional experiments and corresponding findings.
    \item \cref{supp:failure}: Failure cases where the generation quality does not match that of the teacher model.
    \item \cref{supp:mode_seeking}: Visualization and example of the mode-seeking/covering behaviors of the generalized Jeffrey divergence.
    \item \cref{supp:more_visual}: Additional visual results of one-step generations from our distilled models.
    \item \cref{supp:Prompts}: List of all prompts used in this paper for image generation.
\end{itemize}

\section{Relevant Derivations to the Token-level Distillation Loss}
\label{supp:derivation}

\subsection{Loss Gradient}
\label{supp:loss_grad}
Given \cref{eq:softmax} and \cref{eq:on_policy}, by assuming $D$ is differentiable with respect to $p_\theta({x_0^i}|\tilde{x}_t)$, we can calculate \cref{eq:D-grad} using chain rule as: 
\begin{equation}\label{eq:D-grad_supp}
\begin{aligned}
\nabla_\theta\mathcal{L}_\text{\method{}} & = 
   \nabla_{{\theta}} \mathbb{E}_{x_{\text{init}}, t}\!\left[w(t)\!\left(\mathbb{E}_{q_{t|0}}\left[{D\big((p_\phi||p_\theta)(\tilde{x}_t)\big) }\right] \!\right) \right]\! \\
 & = \mathbb{E}_{x_{\text{init}}, t}\!\left[w(t)\!\left(\mathbb{E}_{q_{t|0}}\left[{\nabla_{{\theta}} D\big((p_\phi||p_\theta)(\tilde{x}_t)\big) }\right] \!\right) \right]\!\\
 &=\mathbb{E}_{x_{\text{init}}, t}\!\left[w(t)\!\left(\mathbb{E}_{q_{t|0}}\left[{ {\frac{1}{L_M} \sum_{\substack{i=1 \\ {\tilde{x}_t^i} = \mask{}}}^L 
 \nabla_{{\theta}}D\big(p_\phi({x_0^i}|\tilde{x}_t) \| p_\theta({x_0^i}|\tilde{x}_t)\big)}}\right] \!\right) \right]\! \\
 &=\mathbb{E}_{x_{\text{init}}, t}\!\left[w(t)\!\left(\mathbb{E}_{q_{t|0}}\left[{ {\frac{1}{L_M} \sum_{\substack{i=1 \\ {\tilde{x}_t^i} = \mask{}}}^L \nabla_{{p_\theta({x_0^i}|\tilde{x}_t)}}D\big(p_\phi({x_0^i}|\tilde{x}_t) \| p_\theta({x_0^i}|\tilde{x}_t)\big)}}\frac{\text{d}{p_\theta({x_0^i}|\tilde{x}_t)}}{\text{d}z_\theta^i}\frac{\text{d}z_\theta^i}{\text{d}\theta}\right] \!\right) \right]\!. \\
\end{aligned}
\end{equation}

By defining $\nabla_{{z_\theta}}D\big((p_\phi||p_\theta)(\tilde{x}_t)\big)$ as a vector with the $i$-th element $[\nabla_{{z_\theta}}D\big((p_\phi||p_\theta)(\tilde{x}_t)\big)]_i = \frac{1}{L_M}\nabla_{{p_\theta({x_0^i}|\tilde{x}_t)}}D\big(p_\phi({x_0^i}|\tilde{x}_t) \| p_\theta({x_0^i}|\tilde{x}_t)\big)\frac{\text{d}{p_\theta({x_0^i}|\tilde{x}_t)}}{\text{d}z_\theta^i}$,
and $\frac{\text{d}z_\theta}{\text{d}\theta}$ as a vector with the $i$-th element $\frac{\text{d}z_\theta^i}{\text{d}\theta}$, we have:
\begin{equation}
\begin{aligned}
\nabla&_\theta\mathcal{L}_\text{\method{}} \!=\!
    \mathbb{E}_{x_{\text{init}}, t}\!\left[w(t)\!\left(\mathbb{E}_{q_{t|0}}\!\left[\colororange{\nabla_{{z_\theta}} {{D}(p_\phi||p_\theta)(\tilde{x}_t)}}\,\colorblue{\frac{\text{d} {z_\theta}(\tilde{x}_t)}{\text{d} \theta}}\right] \!\right) \right]\!.
\end{aligned}
\end{equation}
By default, we apply stop-gradient to the $\tilde{x}_t$, since the sample operation from $z_\theta$ to $x_\theta(x_{\text{init}})$ is non-differentiable. 

\subsection{Explicit Form of Divergence}
\label{supp:div_grad}
We start the derivation with the FKL and RKL in the generalized Jeffrey divergence.
The forward and reverse KL between the teacher $\phi$ and the one-step generator $\theta$ at each output location $i$ are:

\begin{equation}\label{eq:fkl_supp}
\begin{aligned}
{D}_{FKL_i}(\tilde{x}_t) &= \sum_{k=1}^{V} p_\phi(x_0^i=k|\tilde{x}_t) \log\left(\frac{p_\phi(x_0^i=k|\tilde{x}_t)}{p_\theta(x_0^i=k|\tilde{x}_t)}\right), \\
{D}_{RKL_i}(\tilde{x}_t) &= \sum_{k=1}^{V} p_\theta(x_0^i=k|\tilde{x}_t) \log\left(\frac{p_\theta(x_0^i=k|\tilde{x}_t)}{p_\phi(x_0^i=k|\tilde{x}_t)}\right).
\end{aligned}
\end{equation}

The derivation of these KLs with respect to the student parameter $\theta$ can be written as:
\begin{equation}\label{eq:fkl_grad1}
\begin{aligned}
\nabla_{\theta}{D}_{FKL_i} &= \sum_{j=1}^{V}\frac{\partial {D}_{FKL_i}}{\partial {z_\theta}^i_j} \frac{\partial {z_\theta}^i_j}{\partial \theta},\\
\nabla_{\theta}{D}_{RKL_i} &= \sum_{j=1}^{V}\frac{\partial {D}_{RKL_i}}{\partial {z_\theta}^i_j} \frac{\partial {z_\theta}^i_j}{\partial \theta}.
\end{aligned}
\end{equation}

Given that the probability corresponding to the logits $z$ as $p_\theta(x_0^i=k|\tilde{x}_t) = \frac{\exp({{z_\theta}^i_k})}{\sum_{n=1}^V \exp({{z_\theta}_n^i})}$, we can precalculate the following quantity for $j\neq k$:
\begin{equation}\label{eq:fkl_supp2}
\begin{aligned}
\frac{\partial}{\partial {{z_\theta}^i_j}} p_\theta(x_0^i=j|\tilde{x}_t) &= \frac{\exp{({z_\theta}^i_j)}}{\sum_{n=1}^V \exp{({z_\theta}^i_n)}}-p_\theta(x_0^i=j|\tilde{x}_t)^2 = p_\theta(x_0^i=j|\tilde{x}_t)(1-p_\theta(x_0^i=j|\tilde{x}_t)), \\
\frac{\partial}{\partial {{z_\theta}^i_j}}p_\theta(x_0^i=k|\tilde{x}_t) &= -\frac{\exp{({z_\theta}^i_k)}\exp{({z_\theta}^i_j)}}{(\sum_{n=1}^V \exp{({z_\theta}^i_n)})^2} = -p_\theta(x_0^i=k|\tilde{x}_t)p_\theta(x_0^i=j|\tilde{x}_t).
\end{aligned}
\end{equation}
These are the gradients of the softmax function that we will use below to help the derivation.

\subsubsection{Gradient of FKL}

For each possible token $i$, we have:
\begin{equation}\label{eq:fkl_grad2}
\small
\begin{aligned}
\frac{\partial {D}_{FKL_i}}{\partial {z_\theta}^i_j} =& \frac{\partial}{\partial {{z_\theta}^i_j}}\sum_{k=1}^{V} p_\phi(x_0^i=k|\tilde{x}_t) \log\left(\frac{p_\phi(x_0^i=k|\tilde{x}_t)}{p_\theta(x_0^i=k|\tilde{x}_t)}\right)\\
=& \frac{\partial}{\partial {{z_\theta}^i_j}}\sum_{k=1}^{V} -p_\phi(x_0^i=k|\tilde{x}_t) \log{p_\theta(x_0^i=k|\tilde{x}_t)} \textcolor[rgb]{0.40,0.40,0.40}{\,\, // \text{From KL to cross entropy}}\\
=& \frac{\partial}{\partial {{z_\theta}^i_j}}\left(\sum_{k=1,k\neq j}^{V} -p_\phi(x_0^i=k|\tilde{x}_t) \log{p_\theta(x_0^i=k|\tilde{x}_t)}\right) - \frac{\partial}{\partial {{z_\theta}^i_j}}p_\phi(x_0^i=j|\tilde{x}_t) \log{p_\theta(x_0^i=j|\tilde{x}_t)} \\
=& \left(\sum_{k=1,k\neq j}^{V} -\frac{p_\phi(x_0^i=k|\tilde{x}_t)}{p_\theta(x_0^i=k|\tilde{x}_t)} \frac{\partial}{\partial {{z_\theta}^i_j}}p_\theta(x_0^i=k|\tilde{x}_t) \right) - \frac{p_\phi(x_0^i=j|\tilde{x}_t)}{p_\theta(x_0^i=j|\tilde{x}_t)} \frac{\partial}{\partial {{z_\theta}^i_j}}p_\theta(x_0^i=j|\tilde{x}_t) \textcolor[rgb]{0.40,0.40,0.40}{\,\, // \text{derivative of } \log(\cdot)}\\
=& \sum_{k=1,k\neq j}^{V} {p_\phi(x_0^i=k|\tilde{x}_t)} {p_\theta(x_0^i=j|\tilde{x}_t)} - {p_\phi(x_0^i=j|\tilde{x}_t)}(1-{p_\theta(x_0^i=j|\tilde{x}_t)}) \textcolor[rgb]{0.40,0.40,0.40}{\,\, // \text{substitute } \cref{eq:fkl_supp2}} \\
=& \sum_{k=1}^{V} {p_\phi(x_0^i=k|\tilde{x}_t)} {p_\theta(x_0^i=j|\tilde{x}_t)} - {p_\phi(x_0^i=j|\tilde{x}_t)} \\
=& {p_\theta(x_0^i=j|\tilde{x}_t)} - {p_\phi(x_0^i=j|\tilde{x}_t)}. \textcolor[rgb]{0.40,0.40,0.40}{\,\, // \text{probability sums up to 1}}
\end{aligned}
\end{equation}

\subsubsection{Gradient of RKL}

For each possible token $i$, we have:
\begin{equation}\label{eq:rkl_grad2}
\small
\begin{aligned}
\frac{\partial {D}_{RKL_i}}{\partial {z_\theta}^i_j} =& \frac{\partial}{\partial {{z_\theta}^i_j}}\sum_{k=1}^{V} p_\theta(x_0^i=k|\tilde{x}_t) \log\left(\frac{p_\theta(x_0^i=k|\tilde{x}_t)}{p_\phi(x_0^i=k|\tilde{x}_t)}\right) \\
=& \frac{\partial}{\partial {{z_\theta}^i_j}}\left(\sum_{k=1,k\neq j}^{V} p_\theta(x_0^i=k|\tilde{x}_t) \log\left(\frac{p_\theta(x_0^i=k|\tilde{x}_t)}{p_\phi(x_0^i=k|\tilde{x}_t)}\right)\right) + \frac{\partial}{\partial {{z_\theta}^i_j}}\left(p_\theta(x_0^i=j|\tilde{x}_t) \log\left(\frac{p_\theta(x_0^i=j|\tilde{x}_t)}{p_\phi(x_0^i=j|\tilde{x}_t)}\right)\right) \\
=& \sum_{k=1,k\neq j}^{V} \left(-p_\theta(x_0^i=k|\tilde{x}_t)p_\theta(x_0^i=j|\tilde{x}_t) \log\left(\frac{p_\theta(x_0^i=k|\tilde{x}_t)}{p_\phi(x_0^i=k|\tilde{x}_t)}\right) - p_\theta(x_0^i=k|\tilde{x}_t)p_\theta(x_0^i=j|\tilde{x}_t) \right) \\
&+ {p_\theta(x_0^i=j|\tilde{x}_t)}(1-{p_\theta(x_0^i=j|\tilde{x}_t)})\log\left(\frac{p_\theta(x_0^i=j|\tilde{x}_t)}{p_\phi(x_0^i=j|\tilde{x}_t)}\right) + {p_\theta(x_0^i=j|\tilde{x}_t)}(1-{p_\theta(x_0^i=j|\tilde{x}_t)}) \textcolor[rgb]{0.40,0.40,0.40}{\,\, // \text{substitute } \cref{eq:fkl_supp2}} \\
=& \sum_{k=1}^{V} \left(-p_\theta(x_0^i=k|\tilde{x}_t)p_\theta(x_0^i=j|\tilde{x}_t) \log\left(\frac{p_\theta(x_0^i=k|\tilde{x}_t)}{p_\phi(x_0^i=k|\tilde{x}_t)}\right) - p_\theta(x_0^i=k|\tilde{x}_t)p_\theta(x_0^i=j|\tilde{x}_t) \right) \\
&+ p_\theta(x_0^i=j|\tilde{x}_t) \log\left(\frac{p_\theta(x_0^i=j|\tilde{x}_t)}{p_\phi(x_0^i=j|\tilde{x}_t)}\right) + p_\theta(x_0^i=j|\tilde{x}_t)\\
=& \sum_{k=1}^{V} \left(-p_\theta(x_0^i=k|\tilde{x}_t)p_\theta(x_0^i=j|\tilde{x}_t) \log\left(\frac{p_\theta(x_0^i=k|\tilde{x}_t)}{p_\phi(x_0^i=k|\tilde{x}_t)}\right)\right) - \cancel{p_\theta(x_0^i=j|\tilde{x}_t)} \textcolor[rgb]{0.40,0.40,0.40}{\,\, // \text{probability sums up to 1}} \\
& + {p_\theta(x_0^i=j|\tilde{x}_t)} \log\left(\frac{p_\theta(x_0^i=j|\tilde{x}_t)}{p_\phi(x_0^i=j|\tilde{x}_t)}\right) + \cancel{p_\theta(x_0^i=j|\tilde{x}_t)} \\
=& p_\theta(x_0^i=j|\tilde{x}_t) \left(\sum_{k=1}^{V} \left( -p_\theta(x_0^i=k|\tilde{x}_t)\log\left(\frac{p_\theta(x_0^i=k|\tilde{x}_t)}{p_\phi(x_0^i=k|\tilde{x}_t)}\right)\right) + \log\left(\frac{p_\theta(x_0^i=j|\tilde{x}_t)}{p_\phi(x_0^i=j|\tilde{x}_t)}\right) \right) \\
=& p_\theta(x_0^i=j|\tilde{x}_t) \left( \log\left(\frac{p_\theta(x_0^i=j|\tilde{x}_t)}{p_\phi(x_0^i=j|\tilde{x}_t)}\right) - {D}_{RKL_i}\right).\textcolor[rgb]{0.40,0.40,0.40}{\,\, // \text{definition of RKL}}
\end{aligned}
\end{equation}
This is similar to the results derived for AR LLM in \cite{wu2024rethinking}.

As a result, the approximated token-level gradients of FKL and RKL at each masked position $i$ in \cref{eq:mdm_div} can be calculated as follows:
\small
\begin{equation}
\begin{aligned}\label{eq:kl_grad_}
\nabla_{z_\psi} 
 D_{FKL_i}\big(p_\phi({x_0}^i|\tilde{x}_t) \| p_\psi({x_0}^i|\tilde{x}_t)\big) := \nabla_{z_{\psi}}{D}_{FKL_i}(\tilde{x}_t) &= ({p_\psi(x_0^i|\tilde{x}_t)} - {p_\phi(x_0^i|\tilde{x}_t)}), \\
 \nabla_{z_\psi} D_{RKL_i}\big(p_\phi({x_0}^i|\tilde{x}_t) \| p_\psi({x_0}^i|\tilde{x}_t)\big) := \nabla_{z_{\psi}}D_{RKL_i}(\tilde{x}_t) &= p_\psi(x_0^i|\tilde{x}_t) \left( \log\left(\frac{p_\psi(x_0^i|\tilde{x}_t)}{p_\phi(x_0^i|\tilde{x}_t)}\right) - {D}_{RKL_i}(\tilde{x}_t) \right).
\end{aligned}
\end{equation}

\subsubsection{Gradient of $f$-divergence}
Our proposed token-level distillation can be seamlessly extended to general $f$-divergence \cite{csiszar2004information} with the form \cite{renyi1961measures,xu2025one,han2024f}:
\begin{equation}\label{eq:fkl}
\begin{aligned}
{D}_{f_i} = \sum_{k=1}^{V} p_\theta(x_0^i=k|\tilde{x}_t) f\left(\frac{p_\phi(x_0^i=k|\tilde{x}_t)}{p_\theta(x_0^i=k|\tilde{x}_t)}\right).
\end{aligned}
\end{equation}

When the generator function $f$ is differentiable, we can calculate its gradient as:
\begin{equation}\label{eq:fkl_grad}
\begin{aligned}
\frac{\partial {D}_{f_i}}{\partial {z_\theta}^i} =& \frac{\partial}{\partial {{z_\theta}^i_j}}\sum_{k=1}^{V} p_\phi(x_0^i=k|\tilde{x}_t) f\left(\frac{p_\phi(x_0^i=k|\tilde{x}_t)}{p_\theta(x_0^i=k|\tilde{x}_t)}\right)\\
=& \sum_{k=1}^{V} \left(
\frac{\partial p_\theta(x_0^i=k|\tilde{x}_t)}{\partial {{z_\theta}^i_j}} \left[f\left(\frac{p_\phi(x_0^i=k|\tilde{x}_t)}{p_\theta(x_0^i=k|\tilde{x}_t)}\right)-\left(\frac{p_\phi(x_0^i=k|\tilde{x}_t)}{p_\theta(x_0^i=k|\tilde{x}_t)}\right)f'\left(\frac{p_\phi(x_0^i=k|\tilde{x}_t)}{p_\theta(x_0^i=k|\tilde{x}_t)}\right) \right] \right) \\
=& p_\theta(x_0^i=j|\tilde{x}_t) \sum_{k=1}^{V} \left(
 -p_\theta(x_0^i=k|\tilde{x}_t) \left[f\left(\frac{p_\phi(x_0^i=k|\tilde{x}_t)}{p_\theta(x_0^i=k|\tilde{x}_t)}\right)-\left(\frac{p_\phi(x_0^i=k|\tilde{x}_t)}{p_\theta(x_0^i=k|\tilde{x}_t)}\right)f'\left(\frac{p_\phi(x_0^i=k|\tilde{x}_t)}{p_\theta(x_0^i=k|\tilde{x}_t)}\right) \right] \right) \\
&+ p_\theta(x_0^i=j|\tilde{x}_t)  
 \left[f\left(\frac{p_\phi(x_0^i=j|\tilde{x}_t)}{p_\theta(x_0^i=j|\tilde{x}_t)}\right)-\left(\frac{p_\phi(x_0^i=j|\tilde{x}_t)}{p_\theta(x_0^i=j|\tilde{x}_t)}\right)f'\left(\frac{p_\phi(x_0^i=j|\tilde{x}_t)}{p_\theta(x_0^i=j|\tilde{x}_t)}\right) \right].  \\
\end{aligned}
\end{equation}

As shown in\cref{tab:f-divergences}, the generalized Jeffrey divergence belongs to $f$-divergence, with generator function $f(u)=((1-\beta)u-\beta) \log u$.

\begin{table*}[!tb]
\centering
\small
\caption{Summary of some typical $f$-divergences ${D}_f(p\|q)$ together with generator
functions $f$, where $f: (0,\infty) \to \mathbb{R}$ is a convex function satisfying the condition $f(1)=0$.
This table is mainly adapted from \cite{han2024f}.
}
\resizebox{\linewidth}{!}{%
\begin{tabular}{lll}
\toprule
Name & ${D}_f(p\|q)$ & Generator $f(u)$ 
\\ 
\midrule
Forward Kullback-Leibler
& $\int p(x) \log \frac{p(x)}{q(x)} \,\textrm{d}x$
& $u \log u$
\\
Reverse Kullback-Leibler
& $\int q(x) \log \frac{q(x)}{p(x)}\,\textrm{d}x$
& $-\log u$
\\
$\alpha$-divergence ($\alpha \notin \{0,1\}$)
& $\frac{1}{\alpha (\alpha-1)} \int
  \left(q(x) \left[\left(\frac{p(x)}{q(x)}\right)^{1-\alpha}- (1 - \alpha)\left(\frac{p(x)}{q(x)}\right) -\alpha\right]\right)
  \,\textrm{d}x$
& $\frac{1}{\alpha(\alpha-1)} \left(u^{1-\alpha} - (1-\alpha) u - \alpha\right)$
\\
Generalized Jeffrey   
& $\int\left[(1-\beta)p(x)-\beta q(x)\right] \log \left(\frac{p(x)}{q(x)}\right) \,\textrm{d}x$
& $((1-\beta)u-\beta) \log u$
\\
Jensen-Shannon
& $\frac{1}{2} \int p(x) \log \frac{2 p(x)}{p(x)+q(x)}
  + q(x) \log \frac{2 q(x)}{p(x) + q(x)}\,\textrm{d}x$
& $-(u+1) \log \frac{1+u}{2} + u \log u$
\\
Squared Hellinger
& $\int\left(\sqrt{p(x)} - \sqrt{q(x)}\right)^2 \,\textrm{d}x$
& $\left(\sqrt{u}-1\right)^2$
\\
\bottomrule
\end{tabular}
}
\label{tab:f-divergences}
\end{table*}

\section{More Discussion on Related Works}
\label{supp:related}

\newcommand{\nameshort}{\texttt{DD}}
\begin{table}[!t]
\centering
\small
{
\caption{Generative performance on class-conditional ImageNet-256. Results of methods in type AR are taken from the \nameshort{}~\cite{liu2024distilled}. Percentage drop values (relative to the teacher) are shown in parentheses.}
\label{tab:ar_dd}
\vspace{-4pt}
\begin{tabular}{clcccccc}
\toprule
Type & Model & FID ($\downarrow$) & IS ($\uparrow$) & Precision ($\uparrow$) & Recall ($\uparrow$) & \#Para & Step ($\downarrow$) \\
\midrule
AR & VAR-d16~\cite{tian2024visual}       & 4.19  & 230.2 & 0.84 & 0.48 & 310M & 10     \\
AR & VAR-d20~\cite{tian2024visual}       & 3.35  & 301.4 & 0.84 & 0.51 & 600M & 10     \\
AR & VAR-d24~\cite{tian2024visual}       & 2.51  & 312.2 & 0.82 & 0.53 & 1.03B & 10     \\
AR & VAR-d16-\nameshort{}~\cite{liu2024distilled}  & 9.94 (137\%)  & 193.6 (16\%) & 0.80 (5\%)  & 0.37 (23\%)  & 327M & \textbf{1} \\
AR & VAR-d16-\nameshort{}~\cite{liu2024distilled}  & 7.82 (87\%)   & 197.0 (14\%) & 0.80 (5\%)  & 0.41 (15\%)  & 327M & 2        \\
AR & VAR-d20-\nameshort{}~\cite{liu2024distilled}  & 9.55 (185\%)  & 197.2 (35\%) & 0.78 (7\%)  & 0.38 (26\%)  & 635M & \textbf{1} \\
AR & VAR-d20-\nameshort{}~\cite{liu2024distilled}  & 7.33 (119\%)  & 204.5 (32\%) & 0.82 (2\%)  & 0.40 (22\%)  & 635M & 2        \\
AR & VAR-d24-\nameshort{}~\cite{liu2024distilled}  & 8.92 (255\%)  & 202.8 (35\%) & 0.78 (5\%)  & 0.39 (26\%)  & 1.09B & \textbf{1} \\
AR & VAR-d24-\nameshort{}~\cite{liu2024distilled}  & 6.95 (177\%)  & 222.5 (29\%) & 0.83 (-1\%) & 0.43 (19\%)  & 1.09B & 2        \\
\cmidrule{2-8}
AR & LlamaGen-B~\cite{sun2024autoregressive}  & 5.42  & 193.5 & 0.83 & 0.44 & 111M & 256    \\
AR & LlamaGen-L~\cite{sun2024autoregressive}  & 4.11  & 283.5 & 0.85 & 0.48 & 343M & 256    \\
AR & LlamaGen-B-\nameshort{}~\cite{liu2024distilled}   & 15.50 (186\%) & 135.4 (30\%) & 0.76 (8\%)  & 0.26 (41\%)  & 98.3M & \textbf{1} \\
AR & LlamaGen-B-\nameshort{}~\cite{liu2024distilled}   & 11.17 (106\%) & 154.8 (20\%) & 0.80 (4\%)  & 0.31 (30\%)  & 98.3M & 2        \\
AR & LlamaGen-L-\nameshort{}~\cite{liu2024distilled}   & 11.35 (176\%) & 193.6 (32\%) & 0.81 (5\%)  & 0.30 (38\%)  & 326M & \textbf{1} \\
AR & LlamaGen-L-\nameshort{}~\cite{liu2024distilled}   & 7.58 (84\%)  & 237.5 (16\%) & 0.84 (1\%)  & 0.37 (23\%)  & 326M & 2        \\
\midrule
MDM   & MaskGit~\cite{chang2022maskgit}    & {6.60}  & {224.07} & {0.831} & {0.402} & 174M & 16  \\
MDM   &  \cellcolor{gray!20} \method{}    & \cellcolor{gray!20} {6.91} (5\%) & \cellcolor{gray!20} {214.05} (4\%) & \cellcolor{gray!20} {0.828}  (0.4\%) & \cellcolor{gray!20} {0.377}  (6.2\%) & \cellcolor{gray!20} 174M & \cellcolor{gray!20} \textbf{1}  \\
\bottomrule
\end{tabular}
}
\end{table}

Recent research has explored the possibility of converting AR models into discrete diffusion versions~\cite{gong2024scaling}. Given this ongoing work, we plan to investigate the applicability of our approach to AR models. Furthermore, while existing limitations of MDM are well known, an efficient MDM could still serve as a draft model for AR speculative decoding~\cite{christopher2024speculative}.
Diffusion distillation has seen widespread adoption of the GAN objective, either directly for distillation~\cite{wang2022diffusion, xu2024ufogen, sauer2025adversarial} or to enhance performance~\cite{yin2024improved, lin2024sdxl, zhou2024adversarial}. However, these methods are not directly applicable to MDM due to its discrete nature.
In a similar vein, recent works based on SiD~\cite{zhou2024score, zhou2024long, zhou2024adversarial, luo2023diff, luo2025one} aim to minimize the Fisher divergence or a generalized score-based divergence for distillation. However, these approaches require backpropagating the gradient through the teacher model—akin to adversarial training—which is infeasible for MDM due to the non-differentiable sampling operation.
Xu \etal~\cite{xu2025one} recently introduced a general framework for distilling continuous diffusion models using $f$-divergence. 
At the cost of training another additional discriminator, this method successfully extends the DMD framework from RKL to general $f$-divergence by utilizing the output of the discriminator to weight the loss gradient in DMD.
However, their method relies on the assumption that the teacher model and the real data used for training the discriminator obey the same underline distribution (otherwise need to simulate the teacher model to get more accurate synthetic data). The training of additional discriminator in general increases the computational overhead.
Moreover, while effective for continuous models, their method relies on an additional GAN loss to achieve better performance, making it unsuitable for MDMs.
To show the performance of our method, we compare it with \nameshort{} \cite{liu2024distilled}, which distills AR models into one-step or few-step generators. As demonstrated in \cref{tab:ar_dd}, our method yields significantly smaller performance drops compared to the teacher models than \cite{liu2024distilled}.
To summarize, our method successfully address the multi-token prediction challenge pointed out in \cite{song2025ideas,hayakawa2024distillation}, by transfer the stochasticity into the model input and always predict tokens from the correct joint distribution of multiple tokens.

\section{More Experiment Setup}
\label{supp:expr_setup}
For the sampling of teacher model, we use the heuristic parallel sampler by default, with temperature=1.3, schedule mode arccos, and choose token to remask randomly during sampling (as the greedy approach by keeping the most confident tokens leads to degraded generation). In \cref{fig:teacher_with_cfg}, we show the teacher's FID with with 5k generated images under-difference inference steps with different CFG.
This result suggests the limitation of the parallel sampler for test time scaling, as pointed out in \cite{ren2025fast}.
Given these results, we use 16 steps and CFG=2.5 for the teacher model by default in the paper.
In order to add random perturbation to the token embeddings, we fix the embedding layer of all the models during distillation.
During the experiment, we use the same mask schedule for getting the intermediate state $\tilde{x}_t$ from one-step model generation and training the \fakemodel{}.
In addition, we choose the same schedule when training the teacher model. We use the arccos schedule for MaskGit distillation and the cosine schedule for Meissonic distillation, respectively.
We adopt the loss weight from \cite{yin2024one} with $w(t) = \frac{1}{p_\theta(x_0|\tilde{x}_t)-p_\phi(x_0|x_\text{init})}$. Our experiment suggests that this weight can prevent the gradient of the generator from exploding (while the one with 1000 times the gradient value can still generate a regular image).
We use by default mixed precision training with \textit{bf16} and a gradient clipping gradient normalization 1.
We use a constant learning rate scheduler with a linear warmup of 100 steps.
We use the \textit{adam} optimizer with beta1 = 0.9 and beta2 =0.999 and no weight decay for all experiments.
All temperatures for the three models are fixed to 1 during distillation.
Exponential Moving Average (EMA) is applied with a rate of 0.9999 for all experiments.
The codebook sizes for MaskGit and Meissonic are 1024 and 8192, and the sequence length of the latent codes for MaskGit and Meissonic are 1024 and 4096, respectively.
For a sequence length of $L$, a codebook size of $V$, and the number of $\mask{}$ tokens to be replaced $N\approx(1-r_{\text{init}})L$, the possible initial token configure is $\tbinom{L}{N}V^N$.
Given that $L$ and $V$ are usually large numbers, the possible initial token configure are sufficient to be the initial state. This explains why the noise perturbation yield marginal improvement, even though stabilize the training.
The embedding layer in the one-step generator is fixed during distillation to improve the stability.
By default, the FID, precision, recall, density and coverage are all calculated with features extracted from the InceptionV3 network.

\section{More Experimental Results}
\label{supp:more_exps}

\begin{figure*}[ht]
    \centering
    \begin{subfigure}[b]{0.335\textwidth}
        \centering
         \includegraphics[width=\textwidth]{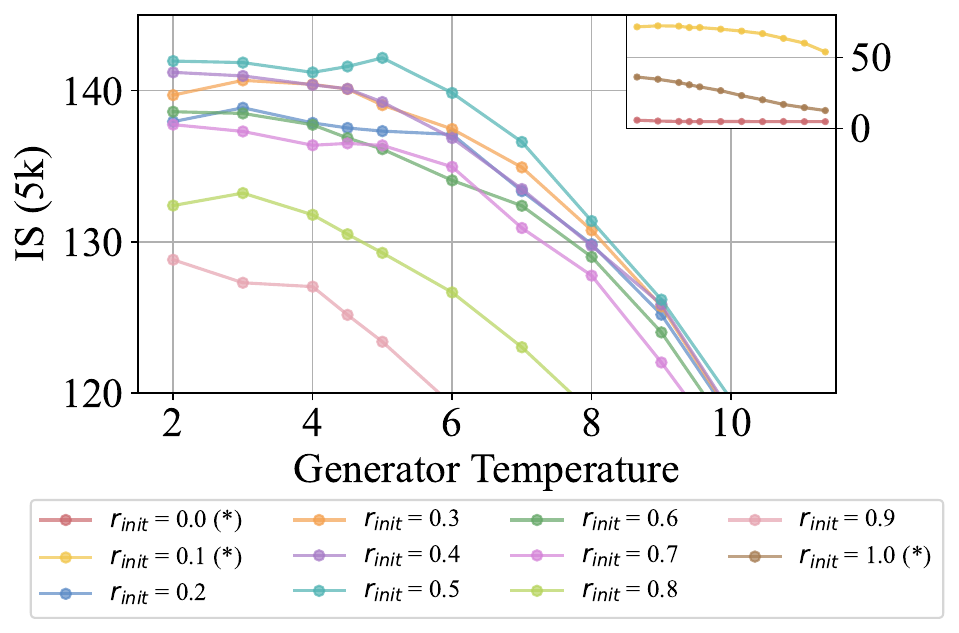}
        \subcaption{{$r_{\text{init}}$}: Initial Mask Ratio}
        \label{subfig:DDMD_Ablation_ratio_IS}
    \end{subfigure}
    \hfill
    \begin{subfigure}[b]{0.32\textwidth}
        \centering
         \includegraphics[width=\textwidth]{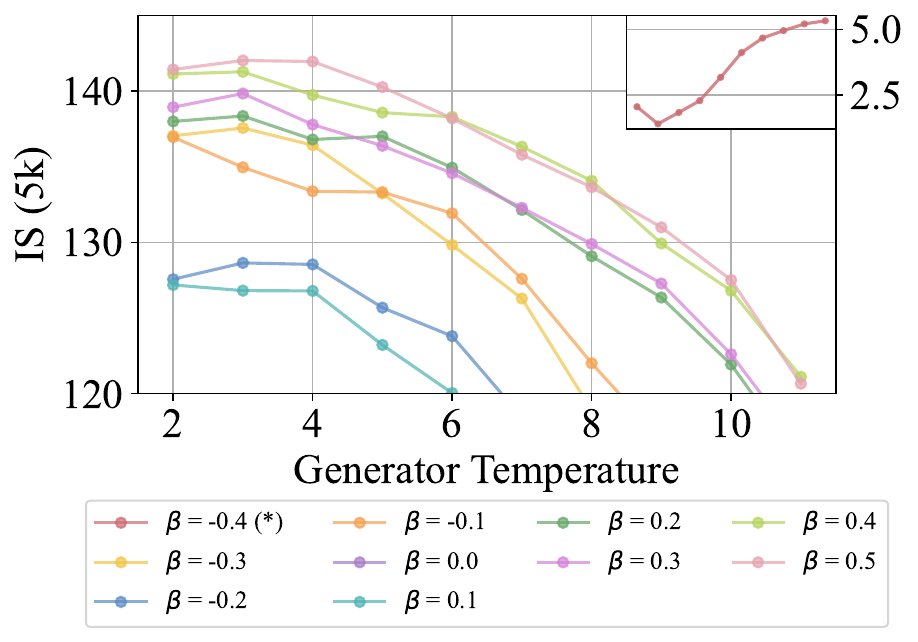}
        \subcaption{\textbf{$\beta$}: Jeffrey Coefficient}
        \label{subfig:DDMD_Ablation_beta_IS}
    \end{subfigure}
    \hfill
    \begin{subfigure}[b]{0.335\textwidth}
        \centering
         \includegraphics[width=\textwidth]{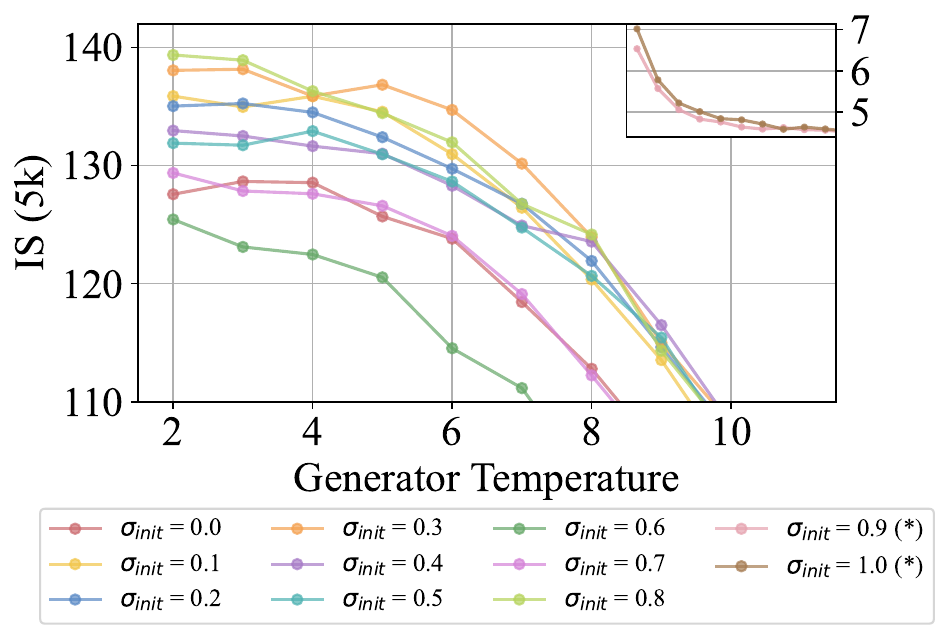}
        \subcaption{$\sigma_{\text{init}}$: Perturbation Strength}
        \label{subfig:DDMD_Ablation_noise_pert_IS}
    \end{subfigure}
    \caption{IS results of the ablations corresponding to \cref{fig:DDMD_Ablation}. $^*$ means {the training is collapsed and falls outside the comparable range with other results}, we therefore show them in the sub-figures at the right upper corner with the same range of the x-axis.}
    \label{fig:DDMD_Ablation_IS}
\end{figure*}

\begin{figure}[h]
\centering
\begin{overpic}[width=0.99\linewidth]{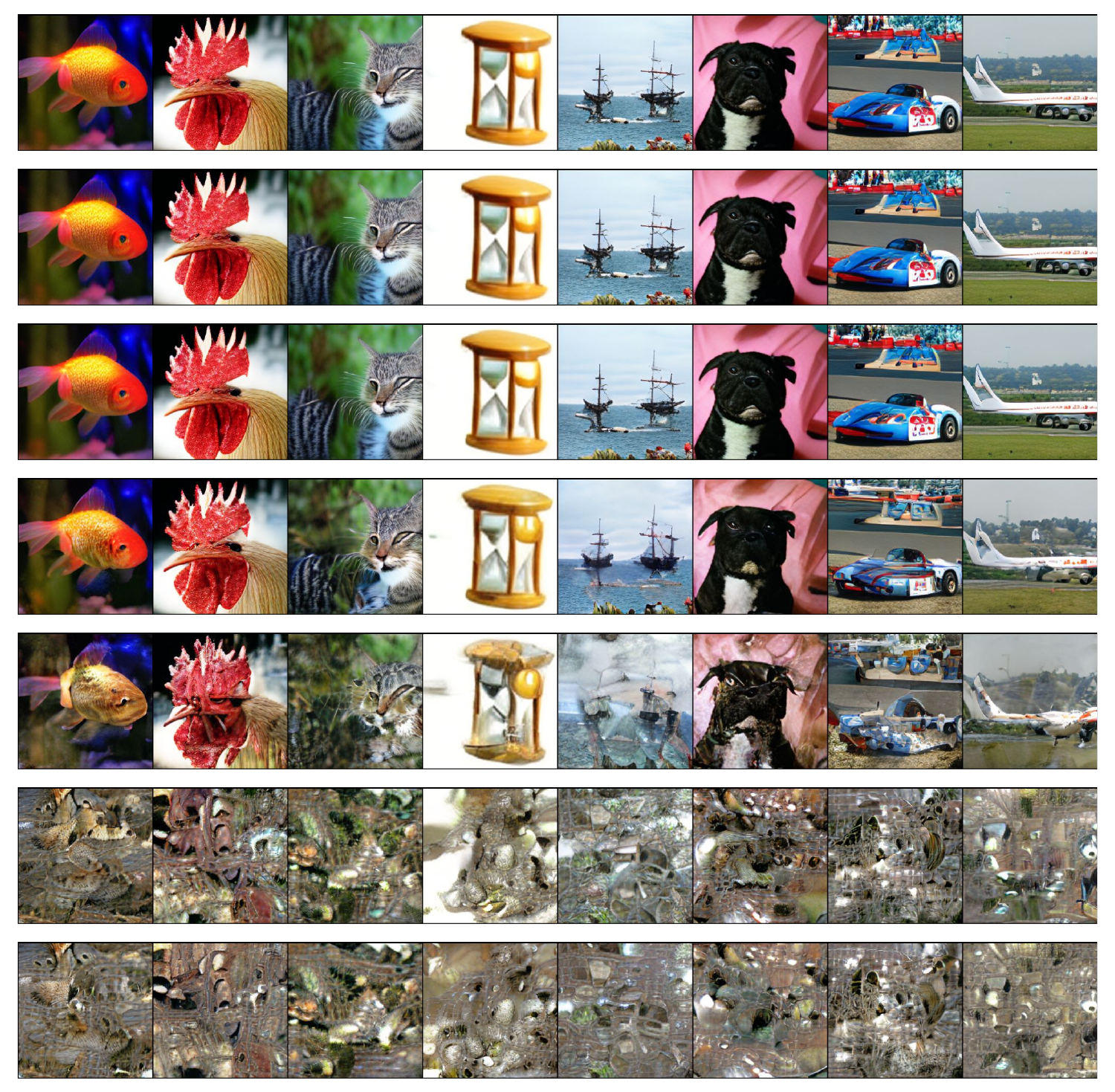}
\end{overpic}
\vspace{-0.2cm}
\caption{One-step generation at different temperature scale.
From top to bottom the temperature is [1e-6,1e-3,1,10,20,50,100]}
\label{fig:temp_scale}
\end{figure}

\begin{figure*}[ht]
    \centering
    \begin{subfigure}[b]{0.505\textwidth}
        \centering
         \includegraphics[width=\textwidth]{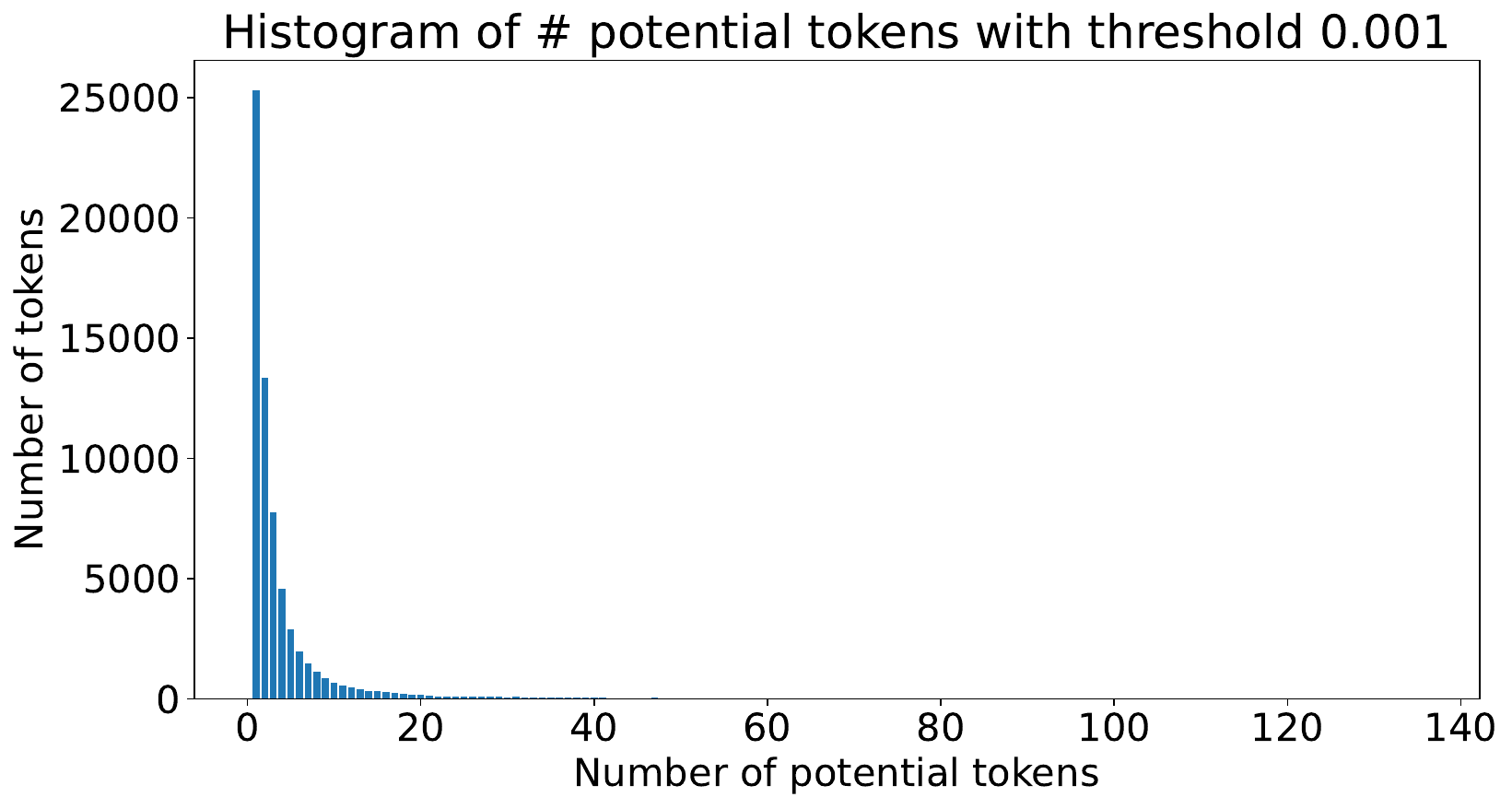}
        \subcaption{Histogram of student probabilities}
        \label{subfig:histogram_distil}
    \end{subfigure}
    \hfill
    \begin{subfigure}[b]{0.49\textwidth}
        \centering
         \includegraphics[width=\textwidth]{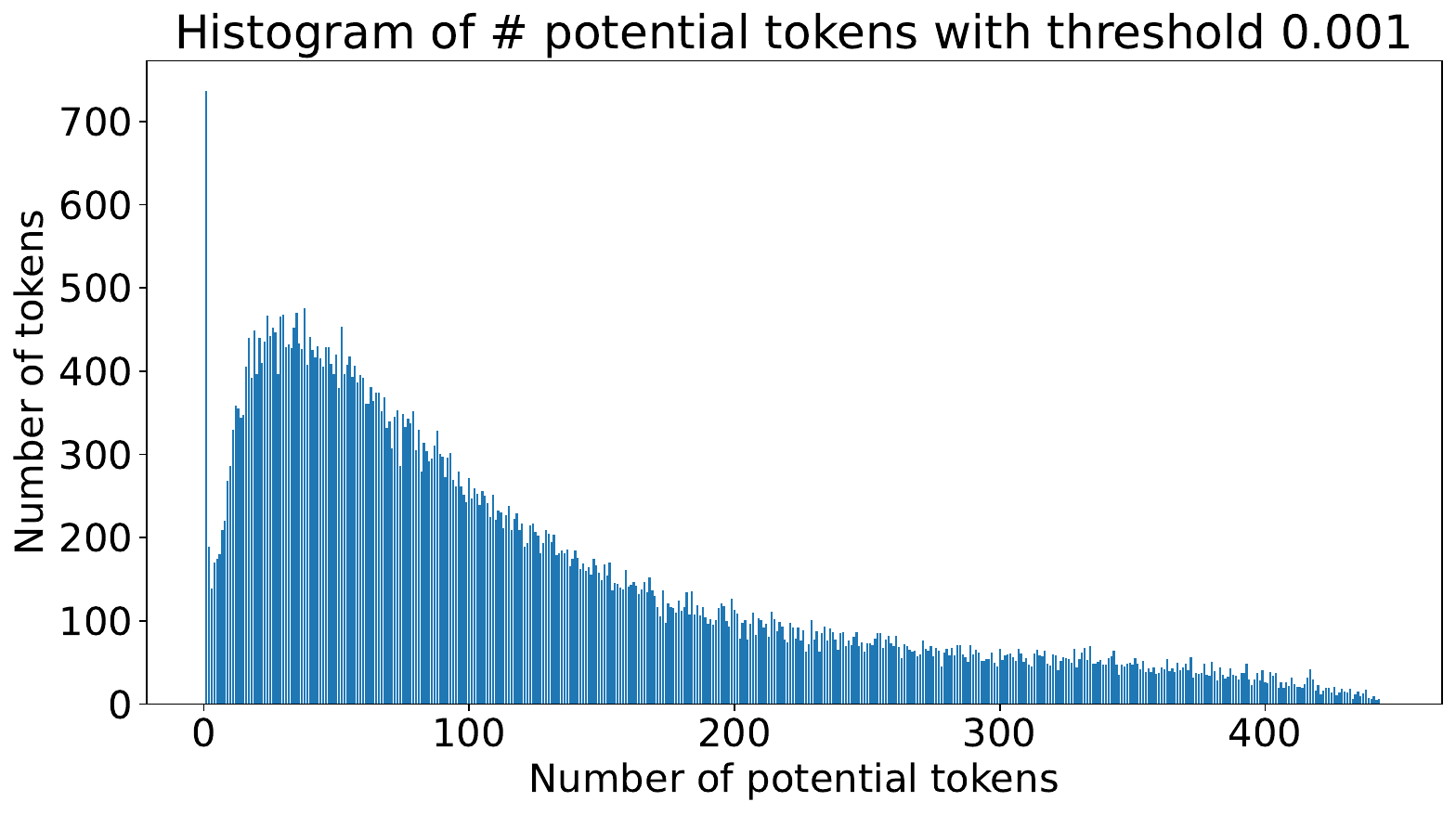}
        \subcaption{Histogram of teacher probabilities}
        \label{subfig:histogram_teacher}
    \end{subfigure}
    \caption{Histogram of number of potential output with probability greater than 0.001}
    \label{fig:histogram}
    \vspace{0.2cm}
\end{figure*}

\begin{figure}[h]
\centering
\begin{overpic}[width=0.99\linewidth]{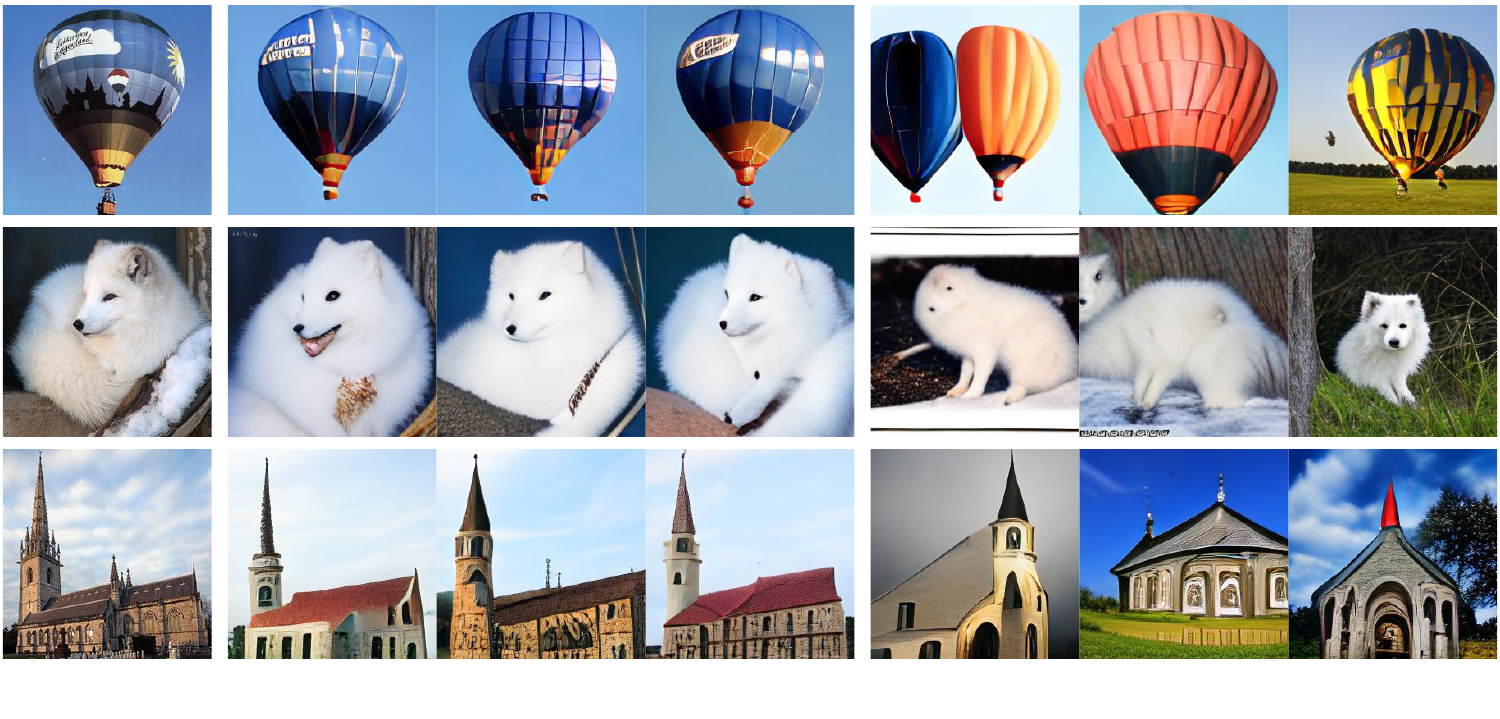}
\put(3.,2.0){\color{black}{\small Real image}}
\put(30.5,1.8){\color{black}{\small Reconstruction}}
\put(72.,1.9){\color{black}{\small{Random generation }}}
\end{overpic}
\vspace{-0.2cm}
\caption{One-step image generation from encoded image tokens and random image tokens. 
The class labels are \texttt{417}, \texttt{279} and \texttt{497}, respectively.}
\label{fig:reconstruction}
\end{figure}
\begin{figure}[h]
\centering
\begin{overpic}[width=0.99\linewidth]{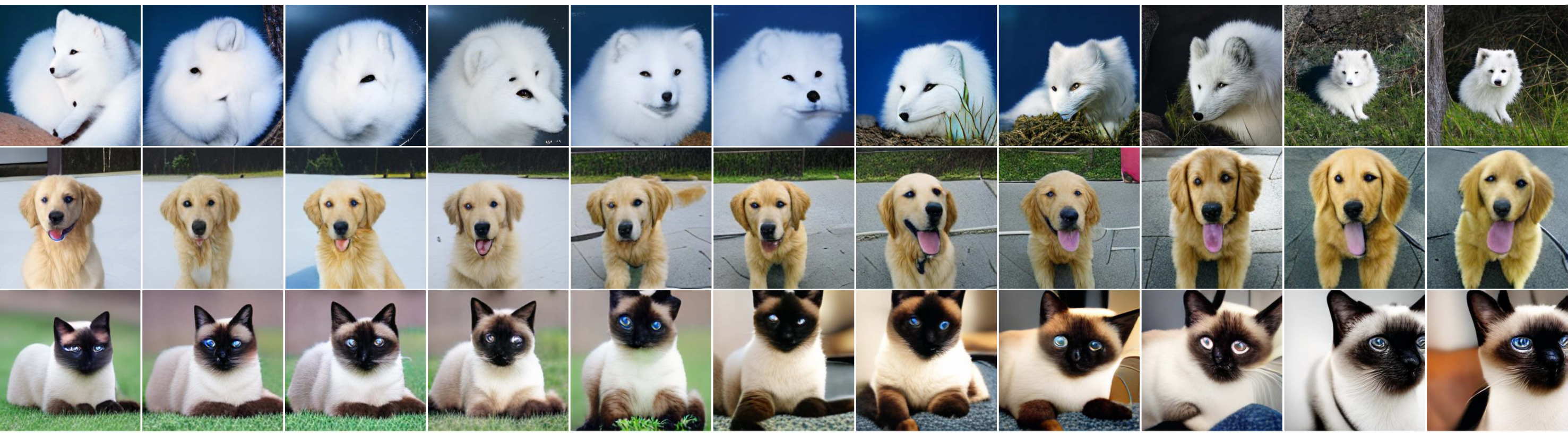}
\end{overpic}
\caption{Interpolation between the encoded image token (left most column) and random image token (right most column)in the initial sequence. 
The class labels are \texttt{279}, \texttt{284} and \texttt{207}, respectively.}
\label{fig:interpolation}
\end{figure}

\begin{table}[!t]
\centering
\caption{
Control experiments on the initialization choices to validate the hypothesis in \cref{sec:init_code}.
}
\resizebox{0.75\linewidth}{!}{
\begin{tabular}{lccc}
\toprule
Initial Method & Step ($\downarrow$)  & FID-5k ($\downarrow$) & IS-5k ($\uparrow$)  \\
\midrule 
use VAE encoded code of random noise & 1 & 189.39 & 4.67 \\
use $1-r_{\text{init}}$ random image tokens, $r_{\text{init}}$ fixed class token (e.g. 1025) & 1 & 173.02 & 5.00 \\
use $1-r_{\text{init}}$ random image tokens, $r_{\text{init}}$ fixed image token (e.g. 512) & 1 & 188.3 & 4.65 \\
use $1-r_{\text{init}}$ random class tokens, $r_{\text{init}}$ fixed mask token $\mask{}$ & 1 & 174.47 & 4.73 \\
\midrule
use $1-r_{\text{init}}$ random image tokens, $r_{\text{init}}$ fixed mask token $\mask{}$ {(ours strategy)} & 1 & 12.01 & 132.44 \\
\bottomrule
\end{tabular}}
\label{table:imagenet_control_init}
\end{table}

In this section we show more experiment results.

\begin{enumerate}

    \item \textbf{IS results from ablation.} We present the IS results of the ablation in \cref{fig:DDMD_Ablation_IS}. Unlike FID, we found that the best IS is achieved at a much lower temperature. This suggest that the temperature can control the trade-off between FID and IS.
    We show visually how the final output image changes with the generator temperature in \cref{fig:temp_scale}.

    \item \textbf{One-step generation metric score with different temperature settings.} In \cref{fig:FID_vs_IS_temp}, we show the FID and IS metric with varying temperatures of the generator. 
    We observe that, as the temperature increases, the IS score decreases, while the FID score initially declines, reaching its minimum at a temperature of 7, before starting to rise again.

    \item \textbf{Top-k visualization.} We apply the top-k trick when sampling each token from the predicted distributions $p_\phi(x_0^i | \tilde{x}_t)$ and $p_\theta(x_0^i | x_\text{init})$. As shown in \cref{fig:top_k}, after distillation, our one-step generation with top-1 sampling and without top-k are nearly identical, unlike in the teacher model. This suggests that each initial code almost deterministically maps to a fixed set of image tokens, justifying the use of Gaussian perturbation for token embeddings in the generator.  
    
    \item \textbf{Analysis of output distribution.} 
    To further investigate this, we examine the number of potential output tokens with probabilities greater than 0.001, as shown in \cref{subfig:histogram_distil}.
    Specifically, we analyze 256 randomly generated images (total 65536 tokens), where each token in every image has a total of 1024 possible output tokens (vocabulary size).
    We find that for nearly half of the tokens, the output logits from our distilled model collapse into a delta distribution, confirming the trends observed in \cref{fig:top_k} and \cref{fig:temp_scale}.
    For comparison, \cref{subfig:histogram_teacher} presents the corresponding distribution for the teacher model, where the potential output token probabilities are more evenly spread across different possible token values.

    \item \textbf{Information in the initial sequence.} Similar to several recent works on the influence of initial noise on the final generated images \cite{chefer2023attend,lugmayr2022repaint,qi2024not,zhou2024golden}. In this work, we designed a token initialization strategy which injects randomness in the initial sequence $x_0$. 
    We here investigate the influence in the initial code. 
    For a distilled model trained with $r_{\text{init}}=0.6$, we tested its performance using initial sequences composed of either 40\% random image tokens or 40\% image tokens derived from encoding real images with VQ-VAE. 
    The results, shown in \cref{fig:reconstruction}, present three image examples with three different random seeds applied to each for both the random image tokens case and the encoded image token case.
    We find that, unlike using random image tokens when the one-step generator can produce diverse images, using the real image tokens instead results in generations that closely resemble the original images.
    This suggests that, similar to continuous diffusion processes, the information contained in randomly initialized codes significantly influences the final generation. This property enables test-time scaling techniques for our model to improve performance \cite{ma2025inference,wang2025remasking}.  
    
    \item \textbf{Interpolation between initial token sequences.} In addition, we show the interpolation results between random initial sequence and encoded sequence in \cref{fig:interpolation}. 

    \item \textbf{Control experiments on token initialization strategy.} In our experiments, we conducted a controlled study on the token initialization strategy to verify our hypothesis in \cref{sec:init_code} that the initial sequence of the student should be similar to those used to train the teacher model (see \cref{table:imagenet_control_init}).
    Similar to our main strategy, we replaced a fraction $1-r_{\text{init}}$ of the $\mask{}$ tokens with random \textit{visual} tokens. However, in this case, we replaced the remaining $\mask{}$ tokens with a fixed token, such as image token 512 or class token 1025.  
    Additionally, we tested another variation where we replaced $1-r_{\text{init}}$ of the $\mask{}$ tokens with random \textit{class} tokens while keeping the rest as $\mask{}$ tokens. 
    In both settings, the models either failed to generate meaningful images.
    This outcome reinforces the student model’s preference for a `familiar' input, a hybrid of visual and $\mask{}$ tokens, during distillation. 
    Furthermore, we experimented with sampling random sequences encoded from Gaussian noise of the same size as an image. As shown in \cref{table:imagenet_control_init}, the training of these control experiments leads to divergence metric values.

    \item \textbf{Additional metrics (FID and CLIP score) on Meissonic.} We evaluated the FID, {Fréchet Dino-v2~\cite{oquab2023dinov2} Distance (FDD)} and CLIP Score \cite{hessel2021clipscore,kang2023scaling} on the MsCoCo 30k validation dataset \cite{lin2014microsoft}, as shown in Table~\ref{tab:FID_CLIP_Meissonic}. Two key observations emerge from the results: (1) as the number of generation steps decreases, the teacher model's FID deteriorates rapidly (e.g., 8-step FID is $96.75$ compared to 64-step FID of $48.27$); and (2) our one-step student achieves superior FID/FDD while maintaining a CLIP Score comparable to the teacher. This suggests that (i) our generator performs competitively with the teacher and {(ii) FID/FDD may not always be a reliable fidelity metric, particularly for the Meissonic teacher model. We hypothesize that the biased style of the teacher model causes discrepancies compared to realistic images from the MS COCO dataset. Consequently, the distilled student model records better performance relative to the teacher.}

    \item \textbf{Additional HPSV2 results on Meissonic.} In \cref{tab:supp_benchmark-hps-v2}, we present a comparison of our teacher model, Meissonic \cite{bai2024meissonic}, across different classifier-free guidance settings. Our results indicate that Meissonic operates optimally at CFG$=9$ for the HPSV2 metric.

    \item We experimented with the Two Time-scale Update Rule \cite{yin2024improved} to update the fake score multiple times per iteration. However, this did not improve model performance or further stabilize the training loss.
    
    \item Inspired by \cite{hinton2015distilling}, we added soft targets for training the \fakemodel{} using the following loss function:
    \begin{align}
    \label{eq:MDM_loss_soft}
        \mathcal{L}_\text{MDM} = \mathbb{E}_{x_{\text{init}}, t}\left[ \gamma(t)\left(\mathbb{E}_{q_{t|0}} [(1-\alpha) (- \log p_{0|t}(x_\theta(x_{\text{init}})|\tilde{x}_t,\phi)) + \alpha D_{\text{KL}}(p_\psi(x_0|\tilde{x}_t)||p_\theta(x_0|x_\text{init})) ] \right)\right] 
    \end{align}
    where $\alpha$ is a hyperparameter controlling the interpolation between the hard target cross-entropy loss and the soft target KL loss.

    \item The default CFG for MaskGit during distillation is 2. We also explored adaptive CFG, where the guidance scale varies with $r_t$, similar to the linear CFG used in MDM parallel sampling. However, this approach did not yield improvements.

    \item Inspired by Proximal Policy Optimization (PPO) \cite{schulman2017proximal}, we introduced an entropy bonus term $-p_\theta(x_0|x_\text{init})\log p_\theta(x_0|x_\text{init})$ to encourage generation diversity. However, this did not show practical benefits.

    \item While $\beta \in [-0.3,1]$ works well for the ImageNet teacher, we found that in the Meissonic experiment, the distillation process diverges when $\beta < -0.1$ or $\beta > 0.2$.

\end{enumerate}

\begin{table}[ht]
  \caption{Complete HPS v2.0 benchmark. Scores are collected from \url{https://github.com/tgxs002/HPSv2}. We highlight the \textbf{best}.}
  \label{tab:supp_benchmark-hps-v2}
  \centering
  \resizebox{.5\linewidth}{!}{
  \begin{tabular}{lcccccc}
    \toprule
     & \multicolumn{2}{r}{\bf{HPS v2.0}}                   \\
    \cmidrule(r){1-7}
    \bf{Model} & \bf{NFE} &\bf{Animation} & \bf{Concept-art} & \bf{Painting} & \bf{Photo} & \bf{Averaged}\\
    \midrule
    Latent Diffusion~\cite{rombach2022high} & 25 & $25.73$ & $25.15$ & $25.25$ & $26.97$ & $25.78$ \\
    DALL·E 2~\citep{ramesh2022hierarchical} & - & $27.34$ & $26.54$ & $26.68$ & $27.24$ & $26.95$ \\
    Stable Diffusion v1.4~\cite{rombach2022high} & 50   & $27.26$ & $26.61$ & $26.66$ & $27.27$ & $26.95$ \\
    Stable Diffusion v2.0~\cite{rombach2022high} & 50  & $27.48$ & $26.89$ & $26.86$ & $27.46$ & $27.17$ \\
    DeepFloyd-XL~\citep{deepfloyddeepfloyd}  & 25  & $27.64$ & $26.83$ & $26.86$ & $27.75$ & $27.27$ \\
    SDXL Base 1.0~\cite{podell2024sdxl}  & 50  & $28.88$ & $27.88$ & $27.92$ & $28.31$ & $28.25$ \\
    SDXL Refiner 1.0~\citep{podell2024sdxl} & 50   & $28.93$ & $27.89$ & $27.90$ & $28.38$ & $28.27$ \\
    InstaFlow \cite{liu2023instaflow} & 1 & 25.98 & 25.79 & 25.93 & 26.32 &  26.01 \\
    SD Turbo \cite{sauer2024adversarial} & 1 & 27.98 & 27.59 & 27.16 & 27.19  & 27.48 \\
    SwiftBrush v2 \cite{dao2025swiftbrush} & 1 & 27.25 & 27.62 & 26.86 & 26.77 & 27.15 \\
    \midrule
    \multirow{6}{*}{Meissonic (cfg=9) \cite{bai2024meissonic}}  & 48 & $\textbf{29.57}$ & $\textbf{28.58}$ & $\textbf{28.72}$ & $\textbf{28.45}$ & $\textbf{28.83}$ \\ %
     & 32  & $29.18$ & $28.32$ & $28.28$ & $27.96$ & $28.44$ \\ 
     & 16  & $28.61$ & $27.82$ & $27.84$ & $27.32$ & $27.90$ \\ 
     & 8  & $25.62$  & $26.49$ & $26.67$ & $27.07$ & $26.46$   \\ 
     & 4  & $25.01$ & $24.95$ & $24.87$ & $23.80$ & $24.66$ \\ 
     & 2  & $23.06$ & $23.28$ & $23.22$ & $22.38$ & $22.98$ \\ 
    \midrule
    \multirow{6}{*}{Meissonic (cfg=4) \cite{bai2024meissonic}}  & 48 &  $28.52$ & $27.44$ & $27.54$ & $27.17$ & $27.67$  \\ %
     & 32  & $28.59$ & $27.54$ & $27.60$ & $27.22$ & $27.74$  \\ 
     & 16  & $28.49$ & $27.52$ & $27.65$ & $27.20$ & $27.71$ \\ 
     & 8  & $27.99$ & $27.24$ & $27.31$ & $26.54$ & $27.27$   \\ 
     & 4  & $26.33$ & $26.03$ & $26.01$ & $24.79$ & $25.79$ \\ 
     & 2  & $23.61$ & $23.87$ & $23.72$ & $22.50$ & $23.43$ \\ 

    \midrule
    \rowcolor{gray!20}
    {\mname{}} & 1  & $28.64$ & $27.91$ & $27.99$ & $27.92$ & $28.11$ \\ 
    \bottomrule
  \end{tabular}
  }
\end{table}

\begin{table*}[]
\small
\centering
\caption{Comparison of FID, FDD and CLIP-Score for Meissonic~\cite{bai2024meissonic} across varying generation steps and our one-step generator. The results are evaluated on MSCOOC-val 30k dataset.}
\begin{tabular}{cccccc}
\toprule
Steps      & 64    & 32    & 16    & 8     & 1 \textit{(ours)} \\
\midrule
FID        & 48.27 & 50.13 & 63.29 & 96.75 & 38.45    \\
FDD        & 620.9 & 625.6 & 709.6 & 980.8 & 548.6    \\
CLIP-Score & 0.321 & 0.318 & 0.307 & 0.280 & 0.322    \\ 
\bottomrule
\end{tabular}
\label{tab:FID_CLIP_Meissonic}
\end{table*}

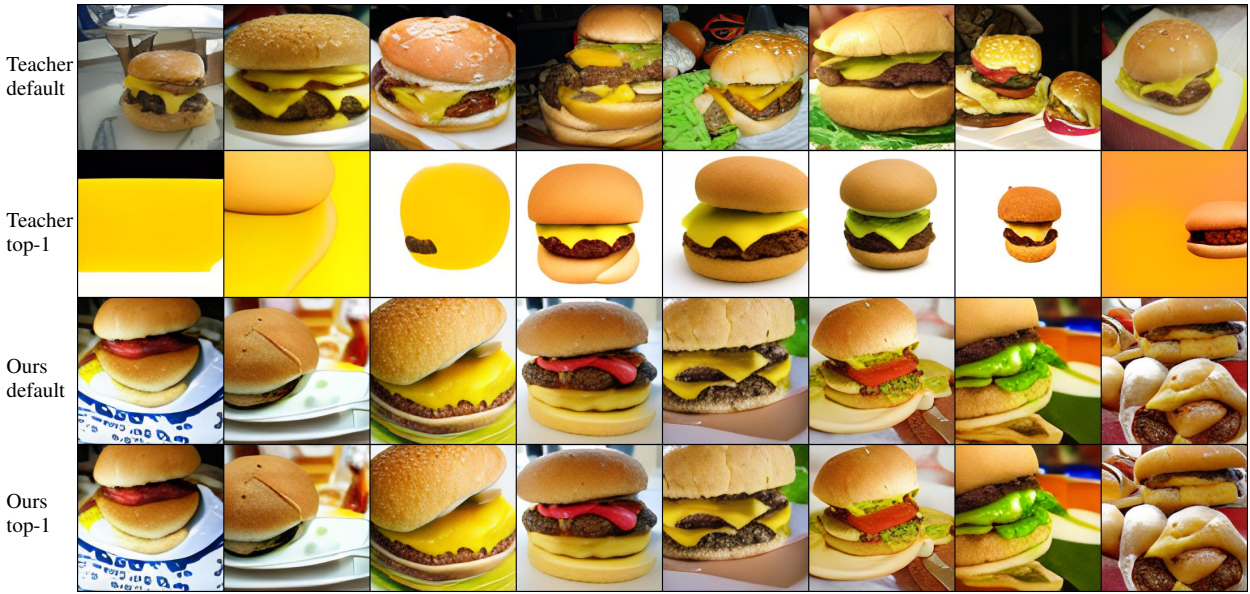
\begin{figure*}[t!]
\centering
\begin{overpic}[width=0.9\linewidth]{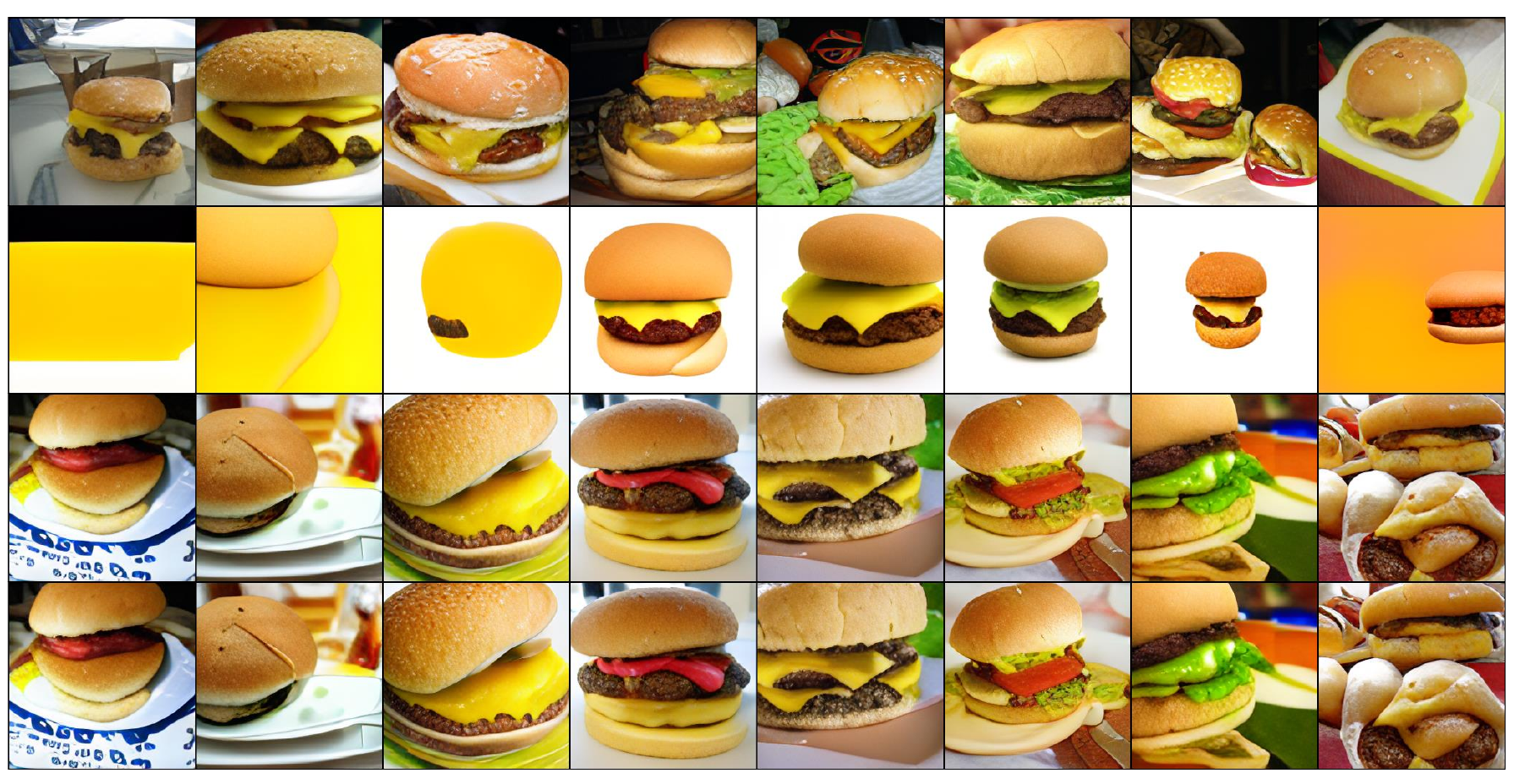}
\put(-5.5,45){\color{black}{\footnotesize Teacher}}
\put(-5.5,43.){\color{black}{\footnotesize default}}
\put(-5.5,31.8){\color{black}{\footnotesize{Teacher }}}
\put(-5.5,29.8){\color{black}{\footnotesize{top-1}}}
\put(-5.5,19.4){\color{black}{\footnotesize{Ours}}}
\put(-5.5,17.4){\color{black}{\footnotesize{default}}}
\put(-5.5,8.0){\color{black}{\footnotesize{Ours}}}
\put(-5.5,6.0){\color{black}{\footnotesize{top-1}}}
\end{overpic}
\vspace{-0.2cm}
\caption{
The top two rows display results from the teacher model using 16-step generation with default setup, with the first row generated without top-k filtering and the other with top-1 sampling.
The bottom two rows showcase results from our distilled one-step generator.
The class label is \texttt{933}.
}
\label{fig:top_k}
\end{figure*}

\section{Failure Cases}
\label{supp:failure}

While the distilled one-step generator performs comparably to the teacher model in most cases, certain classes exhibit slight color shifts and mode collapse, as shown in \cref{fig:mode_color_shift}. This discrepancy may explain the observed gap in evaluation metrics between the teacher and the one-step generator.

\begin{figure*}[t!]
\centering
\begin{overpic}[width=0.9\linewidth]{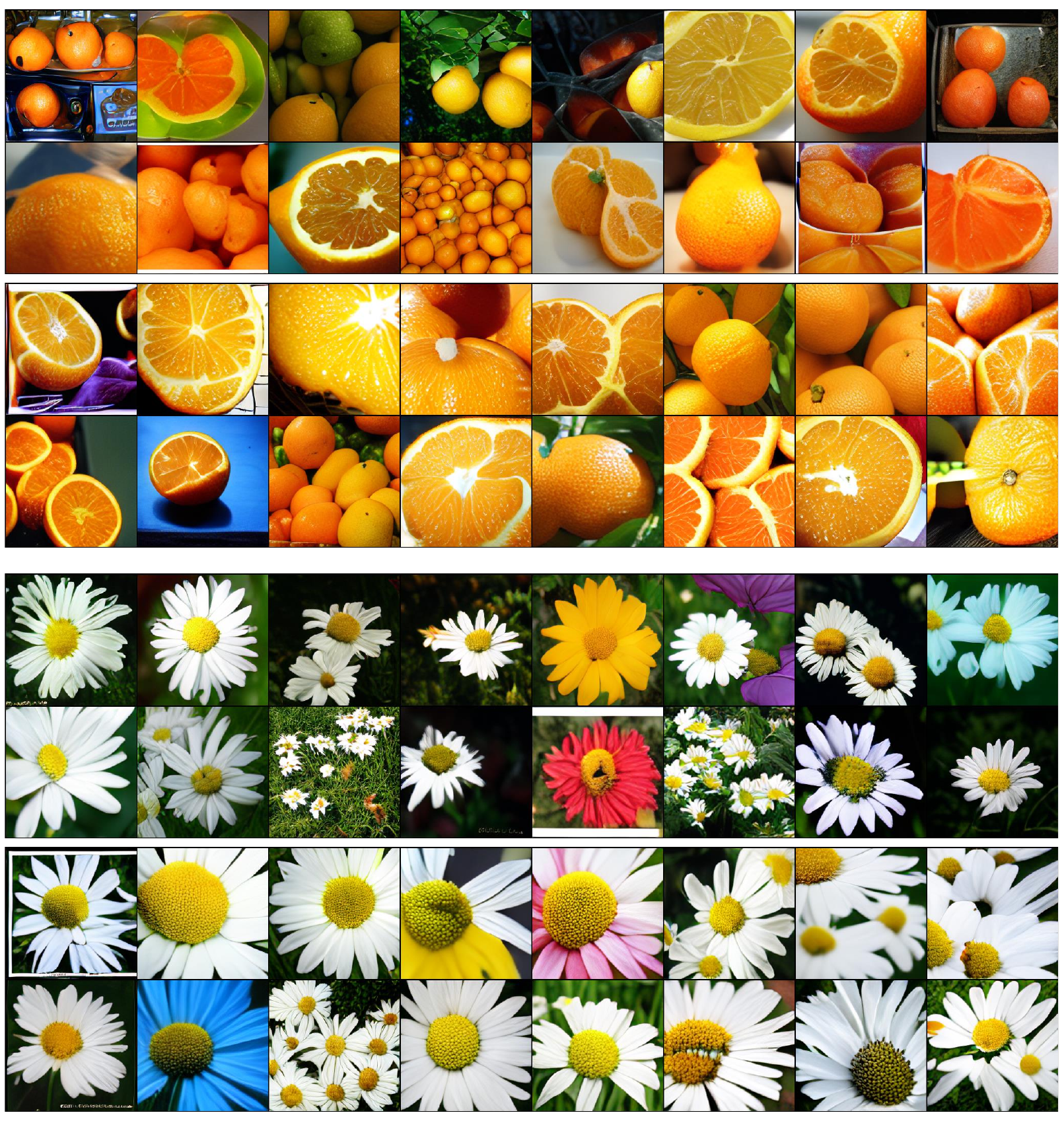}
\end{overpic}
\vspace{-0.2cm}
\caption{
Distribution shift in the student. For both classes, the top two rows are results from the teacher with 16 steps and the lower two rows are the results from our one-step generator.
The class labels are \texttt{950} and \texttt{985}, respectively.
}
\label{fig:mode_color_shift}
\end{figure*}

\begin{figure*}[t!]
    \centering
    \begin{subfigure}[b]{0.49\textwidth}
         \includegraphics[width=\textwidth]{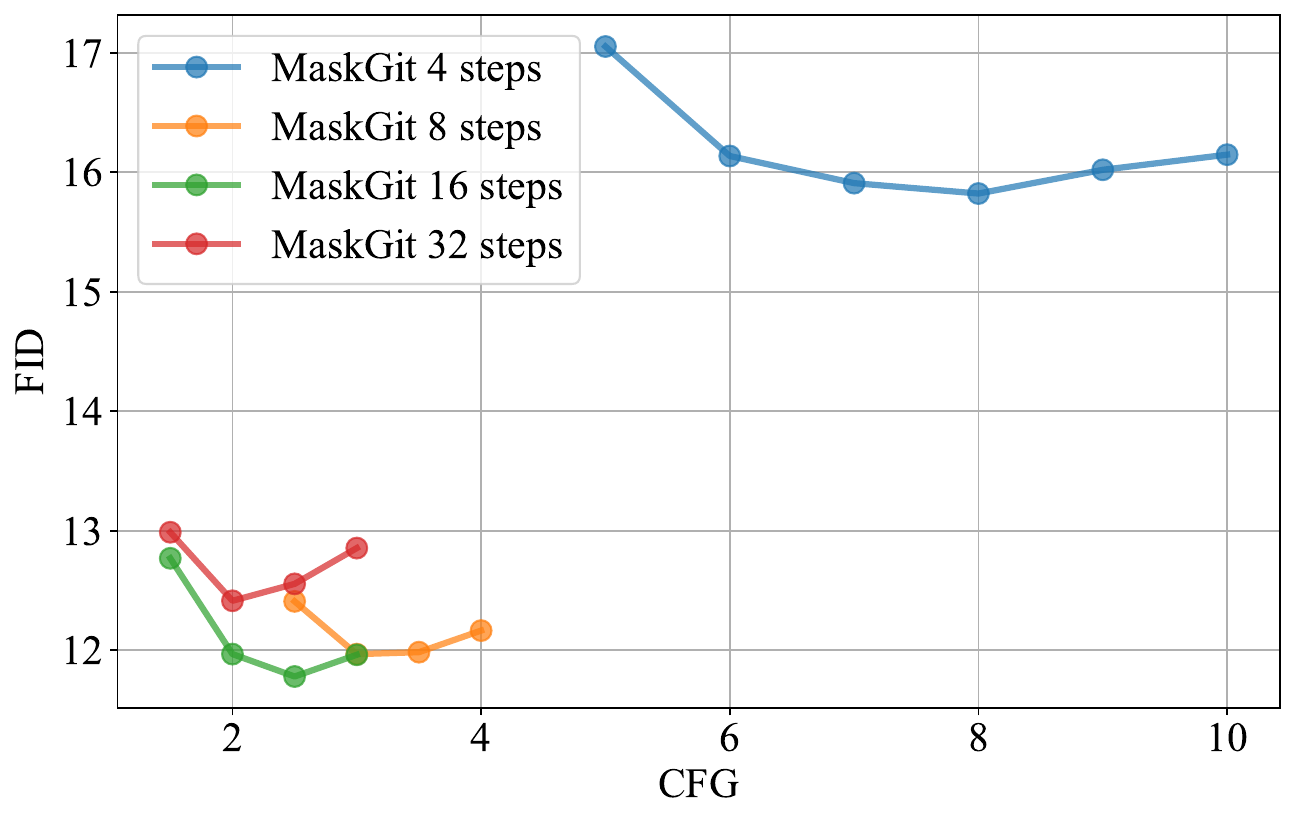}
        \subcaption{MaskGit teacher performance under difference inference steps with different CFG. Metric calculated with 5k generated samples.\\
        \quad \\}
        \label{fig:teacher_with_cfg}
    \end{subfigure}
    \hfill
    \begin{subfigure}[b]{0.49\textwidth}
         \includegraphics[width=\textwidth]{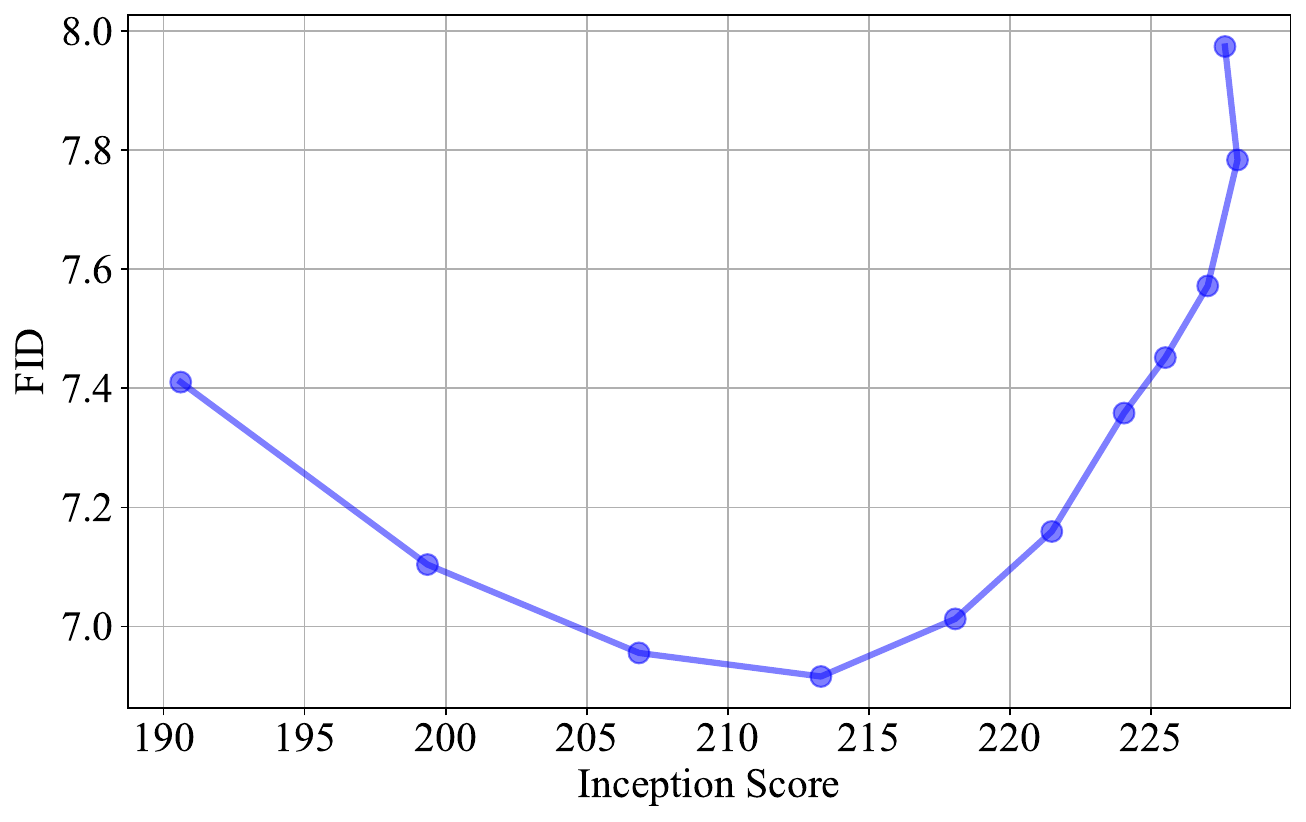}
        \subcaption{Performance of one-step generator distilled from MaskGit teacher under different temperatures. From left to right the temperatures are $[11.0, 10.0, 9.0, 8.0, 7.0, 6.0, 5.0, 4.5, 4.0, 3.0, 2.0]$. Metric calculated with 50k generated samples.}
        \label{fig:FID_vs_IS_temp}
    \end{subfigure}
    \vspace{0.2cm}
\end{figure*}

\begin{figure}[ht!]
\centering
\begin{overpic}[width=0.8\linewidth]{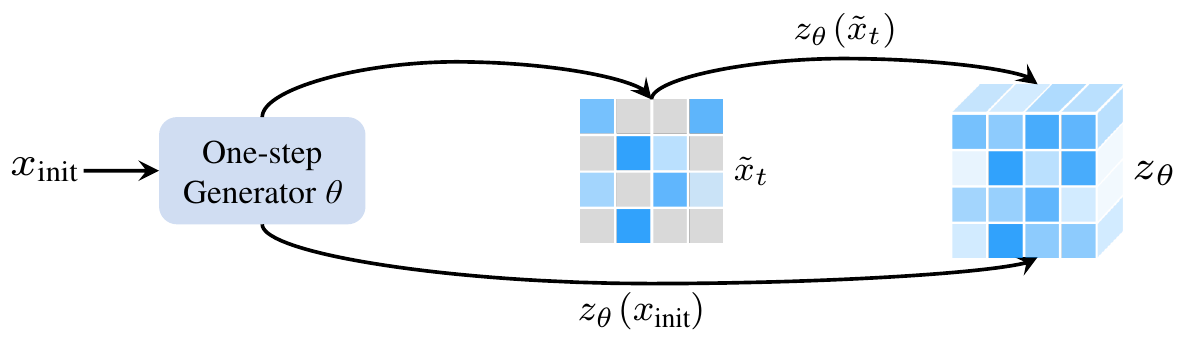}
\end{overpic}
\caption{A visual representation of the consistency assumption. Ideally, the model's prediction based on the correct intermediate state, $\tilde{x}_t$, should be identical to its prediction derived from the initial sequence, $x_{\text{init}}$.}
\label{fig:consistency_assump}
\end{figure}

\section{Mode-seeking vs. Mode-covering}
\label{supp:mode_seeking}
In \cref{fig:jeffrey_div}, we show with a Gaussian example the visualization of the mode-seeking vs. mode-covering behaviors of the generalized Jeffrey divergence with different $\beta$.
When $\beta$ is small, the divergence approaches the FKL, which is known for its mode-covering tendency, assigning high importance to matching all modes of the target distribution. As $\beta$ increases, the behavior transitions toward a balanced form, and for large $\beta$, the divergence exhibits a mode-seeking tendency, akin to the RKL, which focuses on fitting high-density regions while ignoring low-probability modes.

\begin{figure}[ht!]
\centering
\begin{overpic}[width=0.5\linewidth]{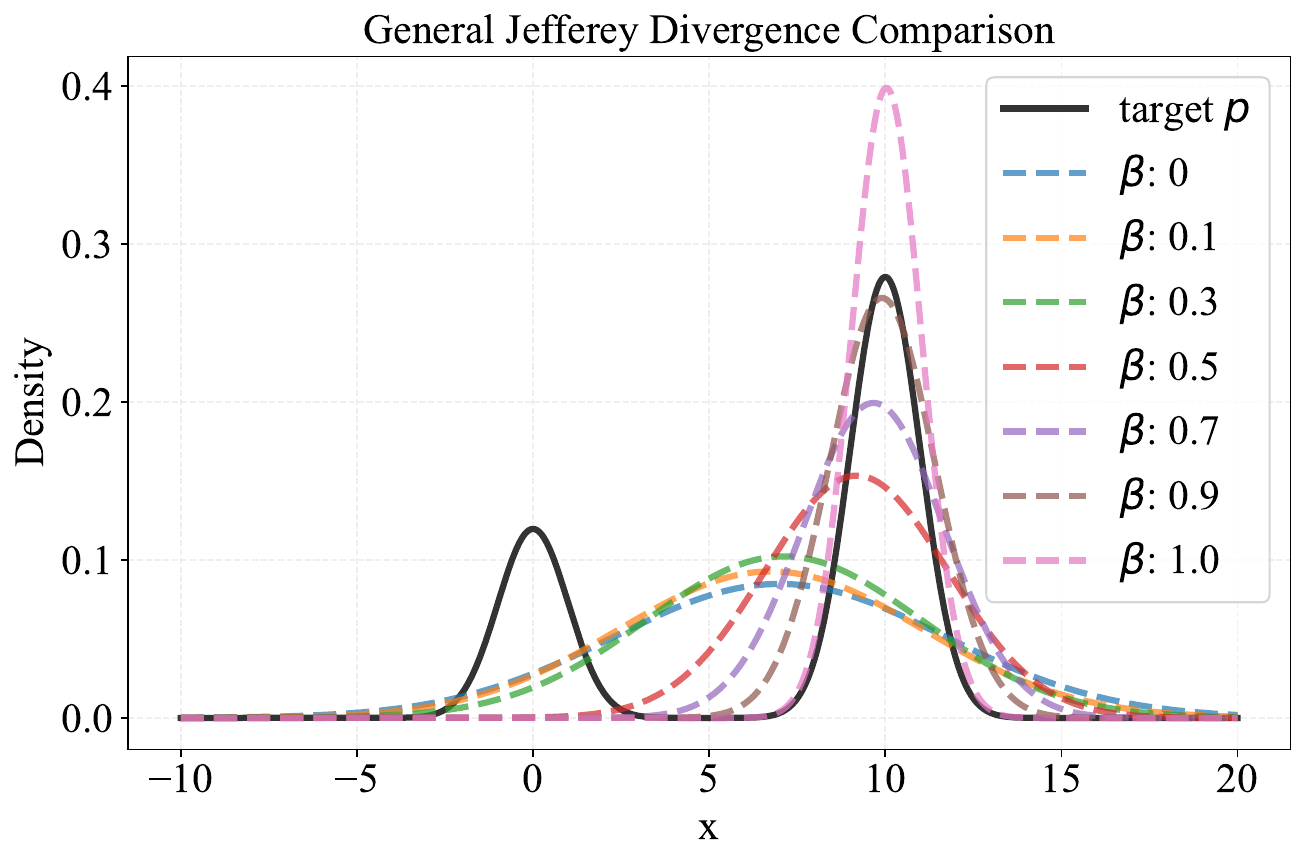}
\end{overpic}
\vspace{-0.2cm}
\caption{Toy example to visualize the mode-seeking VS mode-covering behavior of different $\beta$ values in generalized Jeffrey divergence.
We grid search the mean and std to minimize the generalized Jeffrey divergence.}
\label{fig:jeffrey_div}
\end{figure}

\section{More Qualitative Results}
\label{supp:more_visual}
In \cref{fig:random_imagenet}, we present randomly sampled ImageNet images generated in a single step by our distilled models with MaskGit teacher.
\Cref{fig:comp_teacher} compares our one-step generator with the Meissonic teacher model using different sampling steps. Notably, our one-step generation achieves superior visual quality compared to the teacher model’s 16-step generation.
Finally, in \cref{fig:qulitative_meissonics_sup}, we provide additional text-to-image one-step generation results from our distilled model with the Meissonic teacher.

\begin{figure}[h!]
\centering
\begin{overpic}[width=0.99\linewidth]{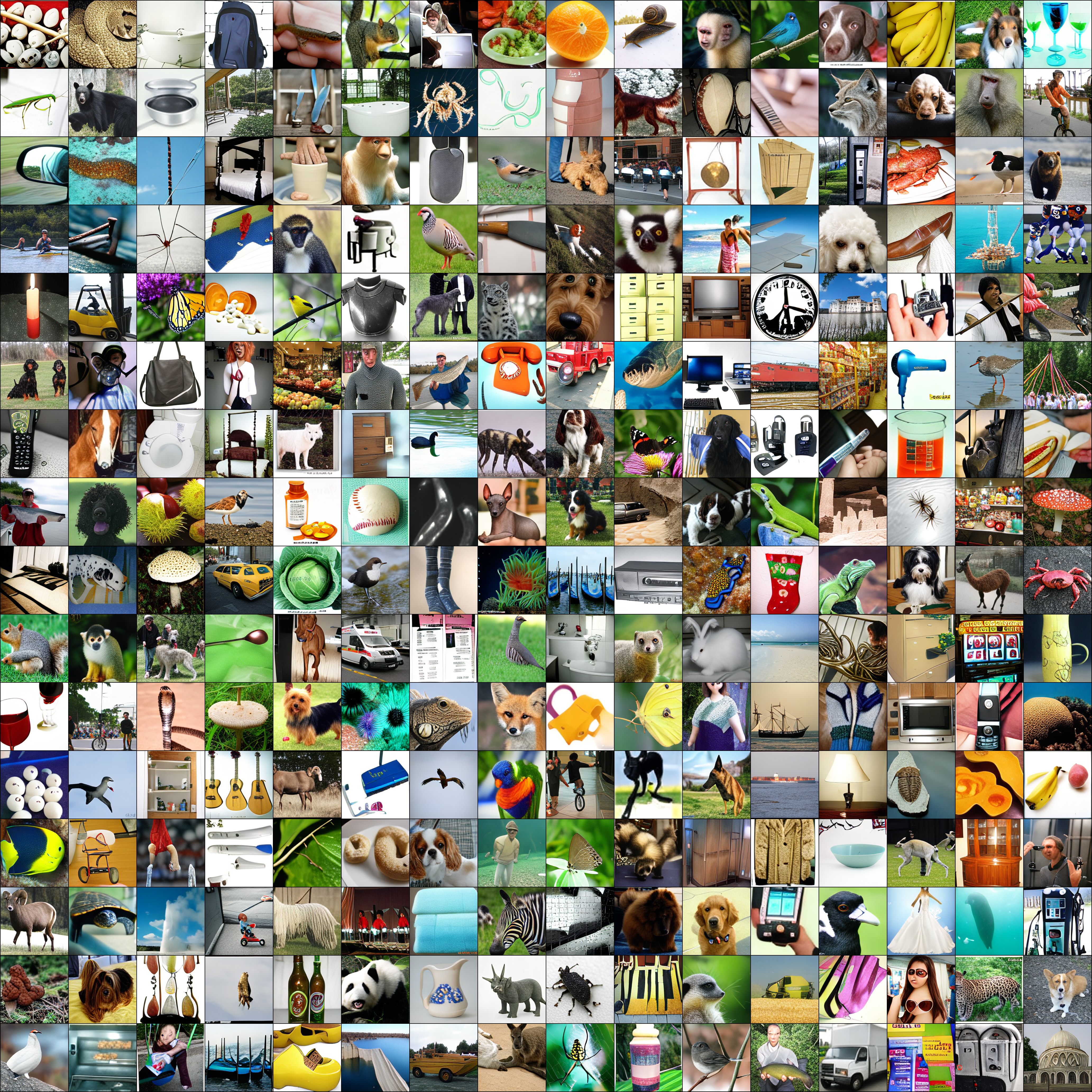}
\end{overpic}
\vspace{-0.2cm}
\caption{One-step samples from our class-conditional model on ImageNet.}
\label{fig:random_imagenet}
\end{figure}

\begin{figure}[h!]
\centering
\begin{overpic}[width=0.99\linewidth]{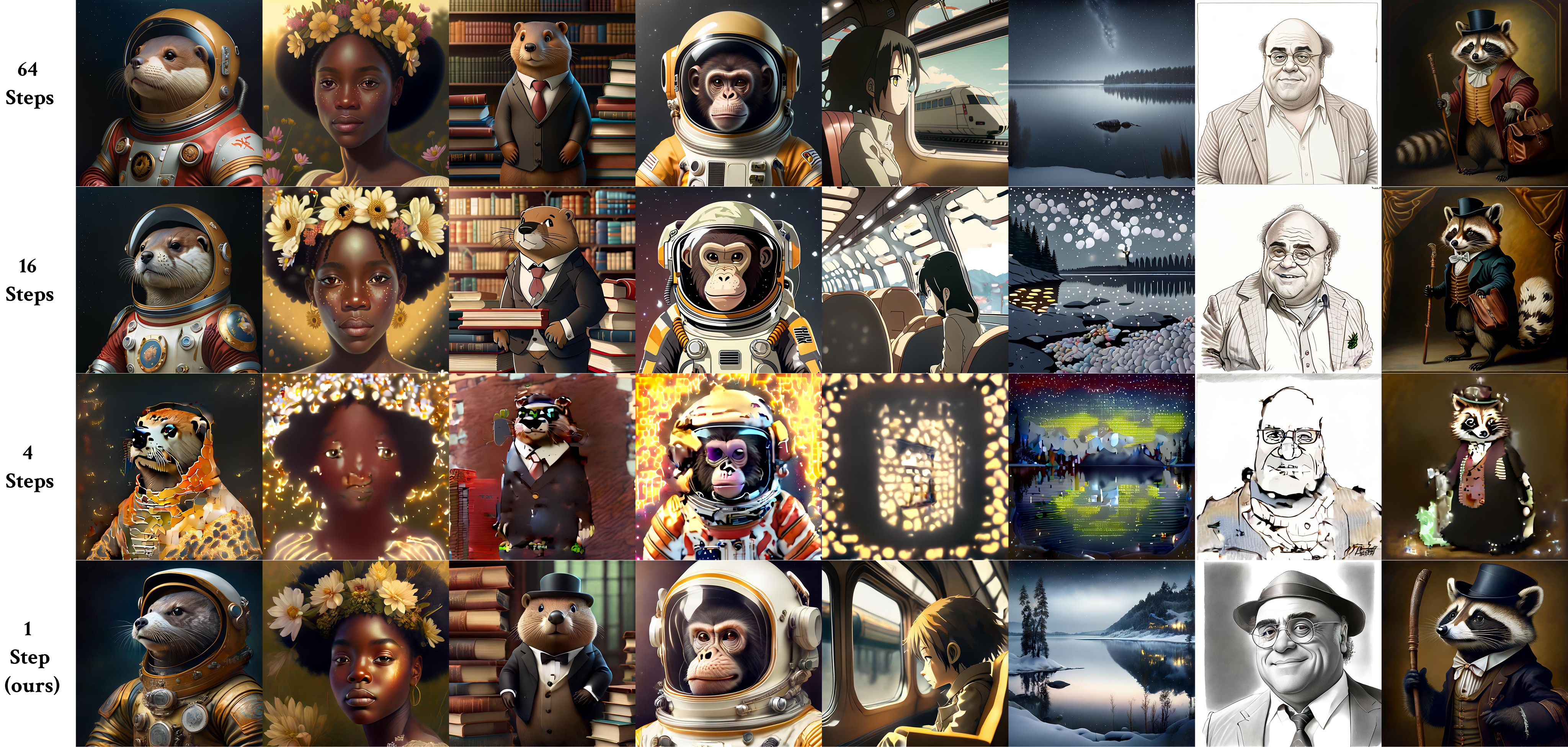}
\end{overpic}
\vspace{-0.2cm}
\caption{Comparison with the teacher: Meissonic~\cite{bai2024meissonic} on different steps, we see clearly that the teacher model's results drop very quickly (e.g., around 4 steps).}
\label{fig:comp_teacher}
\end{figure}

\begin{figure}[h!]
\centering
\begin{overpic}[width=0.99\linewidth]{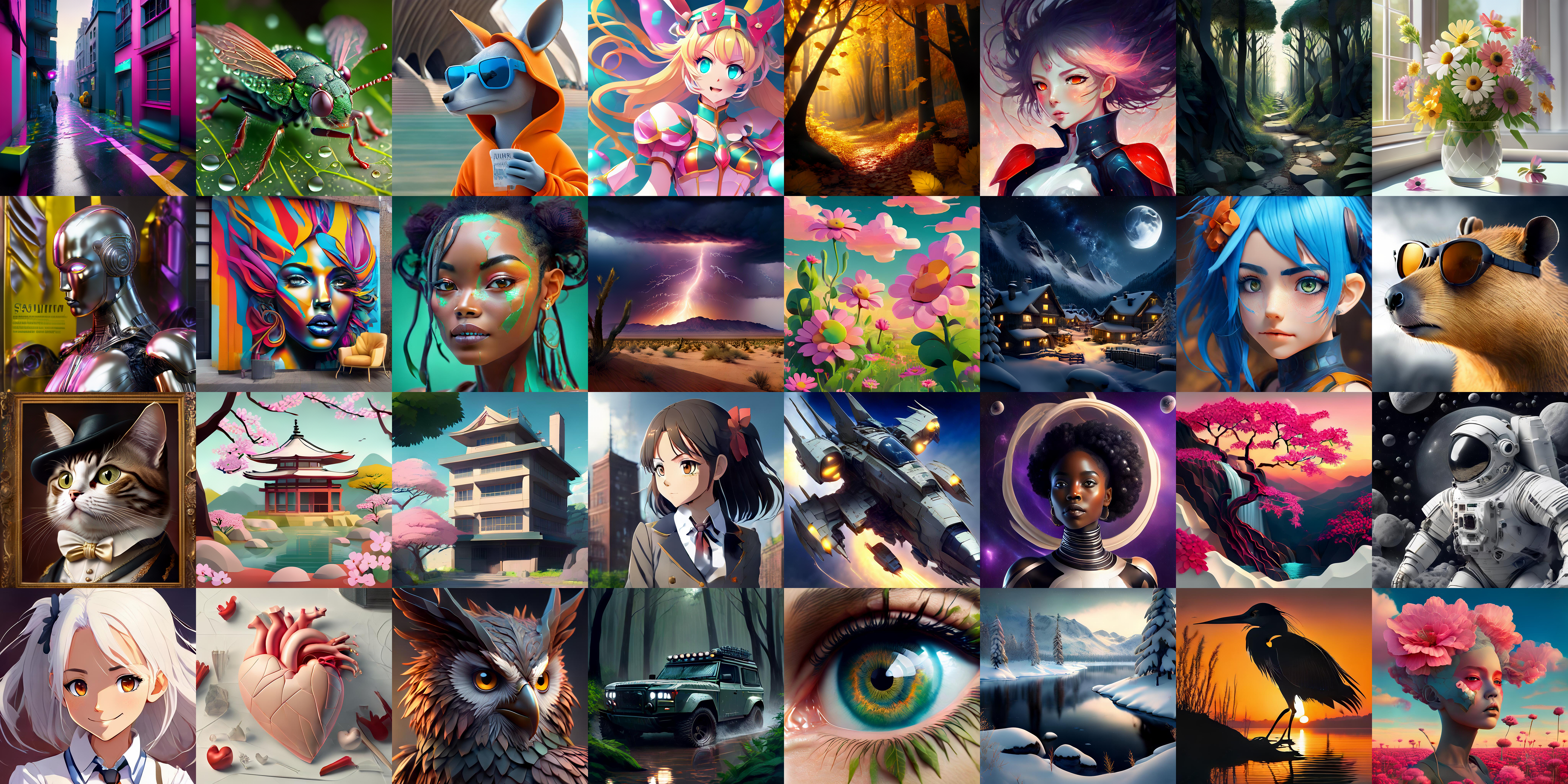}
\end{overpic}
\vspace{-0.2cm}
\caption{Qualitative results of our one-step generator distilled from Meissonic~\cite{bai2024meissonic}.}
\label{fig:qulitative_meissonics_sup}
\end{figure}

\clearpage
\newpage

\section{Misc.}
\label{supp:Prompts}
\paragraph{Prompts}
Below is a collection of creative prompts we used to generate images in \cref{fig:teaser,fig:qualitative_meissonic,fig:qulitative_meissonics_sup,fig:comp_teacher}:
\begin{itemize}
    \item A plushy tired owl sits on a pile of antique books in a humorous illustration.
    \item A photograph of a woman from Steven Universe with gigantic pink ringlets and a white dress.
    \item A white bichon frise puppy dog riding a black motorcycle in Hollywood at sundown with palm trees in the background.
    \item A photorealistic image of a giant floating glass sphere in a rocky landscape surrounded by a gentle mist.
    \item A cosmonaut otter poses for a portrait painted in intricate detail by Rembrandt.
    \item A beaver in formal attire stands next to a stack of books in a library.
    \item An egirl with pink hair and extensive makeup.
    \item Portrait of a monkey wearing a spacesuit and an astronaut helmet.
    \item Closeup of a seinen manga film still showing the interior of a shinkansen train with a leather seat and a window view, with a hyperrealistic film still from a Nepali movie projecting in the background.
    \item Swedish lake at night with heavy snowfall depicted in hyper-realistic and detailed art.
    \item Pencil sketch of Danny DeVito by Milt Kahl.
    \item A realistic anime painting of a cosmic woman wearing clothes made of universes with glowing red eyes.
    \item Photo of Ty Lee from Avatar.
    \item A green field with flowers and pink and yellow clouds under a bright sun at sunset, illustrated by Peter Chan in a colorful Day of the Tentacle style on Artstation.
    \item A raccoon in formal attire, carrying a bag and cane, depicted in a Rembrandt-style oil painting.
    \item A girl in school uniform standing in the city.
    \item Serene, anime-style landscape with vibrant flowers and trees, picturesque clouds, and no signs of human activity.
    \item A kangaroo wearing an orange hoodie and blue sunglasses holding a sign in front of the Sydney Opera House.
    \item A pikachu in a forest illustration.
    \item An oil painting close-up portrait of a young black woman wearing a crown of wildflowers, surrounded by hazy golden light.
    \item A capybara wearing sunglasses.
    \item A frog wearing an anime-inspired onesie.
    \item A blue-haired girl with soft features stares directly at the camera in an extreme close-up Instagram picture.
    \item Digital art of Prince of Roses.
    \item A landscape featuring a Kyoto Animation-style building.
    \item A path winding through a forest depicted in digital art.
    \item A close-up portrait of a beautiful girl with an autumn leaves headdress and melting wax.
    \item A neon-soaked cyberpunk alleyway with rain-drenched streets and futuristic holograms, gritty yet vibrant, hyper-realistic, ultra-detailed, cinematic scene.
    \item A serene mountain landscape at sunrise, mist rolling over rugged peaks, ultra-detailed, photorealistic, soft lighting, high-resolution, digital art.
    \item A hyper-detailed closeup of a dew-covered insect on a vibrant leaf, extreme macro photography style, ultra-realistic, high-resolution, intricate textures.
    \item An anime-style magical girl in a dynamic pose, vibrant colors, ultra-detailed costume and background, energetic, high-resolution, cinematic lighting.
    \item An enchanted autumn forest with falling leaves and warm, glowing light, ultra-detailed, photorealistic, rich textures, digital art, serene mood.
    \item An elegant Renaissance portrait of a noble figure, detailed textures, soft natural lighting, ultra-detailed, classical, high-resolution, oil painting style.
    \item A cybernetic humanoid robot portrait with metallic textures and neon accents, ultra-detailed, photorealistic, cinematic, futuristic digital art.
    \item A vibrant street art mural on an urban wall, ultra-detailed, energetic, bold colors, high-resolution, digital painting, modern art style.
    \item A dark fantasy warrior in intricately detailed armor standing in a stormy battlefield, ultra-detailed, hyper-realistic, cinematic, dynamic action scene.
    \item An intense lightning storm over a vast desert landscape, ultra-detailed, dramatic, high-resolution, cinematic, digital art, atmospheric.
    \item A detailed nature macro shot of a vibrant flower with dewdrops, ultra-detailed, photorealistic, high-resolution, digital painting, delicate textures.
    \item An ultra-realistic snowy mountain village under a starry sky, ultra-detailed, atmospheric, cinematic, high-resolution, digital winter wonderland.
    \item A humorous portrait of a cat dressed as a Victorian aristocrat, in vintage photorealism.
    \item A photorealistic shot of a bouquet of wildflowers in a clear glass vase on a sunlit windowsill.
    \item A mecha jet fighter engages in an air battle with an explosion as a backdrop, set against a dark, starry sky in a highly-detailed art piece by Stephan Martiniere.
    \item A young black woman stands in front of a ringed planet in space.
    \item Digital art of a cherry tree overlooking a valley with a waterfall at sunset.
    \item An astronaut in white futuristic cybernetic armor running on the surface of the moon, featured in an artwork illustration on Artstation.
    \item The image is a headshot of a happy girl with white hair in a school uniform, illustrated by Ilya Kuvshinov.
    \item A minimalistic heart drawing created using Adobe Illustrator.
    \item The image is a digital art headshot of an owlfolk character with high detail and dramatic lighting.
    \item Close up of an eye with the Earth inside the pupil, inspired by Wes Anderson's art.
    \item Landrover drives through a rain-soaked forest in a highly-detailed digital artwork by Greg Rutkowski and Artgerm.
    \item A snowy lake in Sweden captured in a vibrant, cinematic style with intense detail and raytracing technology showcased on Artstation.
    \item A heron silhouetted against a beautiful sunrise, created by Greg Rutkowski.
    \item A surreal portrait of a woman with a giant carnation face in a flower field at sunset with colorful clouds and a large sky, created by artist Simon Stålenhag.
\end{itemize}


\end{document}